\documentclass{article}
\usepackage{iclr2027_conference,times}

\usepackage{amsmath,amsfonts,bm}

\def\eqref#1{equation~\ref{#1}}
\def\1{\bm{1}}

\DeclareMathAlphabet{\mathsfit}{\encodingdefault}{\sfdefault}{m}{sl}
\SetMathAlphabet{\mathsfit}{bold}{\encodingdefault}{\sfdefault}{bx}{n}

\usepackage{titlesec}

\usepackage{titlesec}

\titlespacing*{\section}
  {0pt}            % left
  {2ex plus .3ex minus .2ex}   % before
  {0.7ex plus .2ex}              % after

\titlespacing*{\subsection}
  {0pt}
  {1.2ex plus .3ex minus .2ex}
  {0.5ex plus .2ex}

\titlespacing*{\subsubsection}
  {0pt}
  {1.0ex plus .2ex minus .2ex}
  {0.4ex plus .1ex}

\usepackage{hyperref}
\usepackage{url}
\usepackage{booktabs}
\usepackage{amssymb}
\usepackage{placeins}
\usepackage{array}
\usepackage{colortbl}
\usepackage{arydshln}
\usepackage{graphicx,multirow}
\usepackage[table]{xcolor}

\hypersetup{
    colorlinks=true,
    citecolor=blue,
    linkcolor=black,
    urlcolor=blue
}

\title{EmoGenie: Benchmarking and Evaluating Expressed and Evoked Emotion in T2I Generation with MLLMs}

\author{
\textbf{Zixuan Xing\textsuperscript{1,$\ast$}, Lancheng Gao\textsuperscript{1,$\ast$}, Shengyan Li\textsuperscript{1}, Huiyu Duan\textsuperscript{1},} \\
\textbf{Zhaolin Cai\textsuperscript{1}, Guangtao Zhai\textsuperscript{1}, Xiongkuo Min\textsuperscript{1,$\dagger$}} \\[1ex]
\textsuperscript{1}Shanghai Jiao Tong University 
}

\iclrfinalcopy % Uncomment for camera-ready version, but NOT for submission.
\begin{document}
\maketitle
% No hyperlinks in header

%% Please note that we have introduced automatic line number generation
%% into the style file for \LaTeXe. This is to help reviewers
%% refer to specific lines of the paper when they make their comments. Please do
%% NOT refer to these line numbers in your paper as they will be removed from the
%% style file for the final version of accepted papers.

\begin{abstract}
Rapid advances in artificial intelligence-generated content (AIGC) have significantly promoted using text-to-image (T2I) models to generate images with specific affective intentions. However, the emotional effects conveyed by generated images frequently deviate from target expectations, and it remains unclear whether multimodal large language models (MLLMs) can reliably interpret the emotion of AI-generated visual content. Existing affective benchmarks are predominantly built on natural images, while affective studies of generated images have largely focused on  emotions expressed in portraits, lacking a unified benchmark for studying the emotions expressed and evoked by generated images. We therefore introduce \textbf{EmoGenie}, a benchmark of \underline{emo}tion for AI-\underline{gen}erated \underline{i}mages that \underline{e}valuates capabilities in T2I generation and MLLM understanding across expressed and evoked emotions.   It comprises $6{,}400$ images generated by eight T2I models using $800$ emotion-targeted prompts, annotated with question-answer (QA) pairs, termed \textbf{EmoGenie-QA}, and continuous human ratings, termed \textbf{EmoGenie-Assess}. EmoGenie-QA contains $90{,}000$ closed-set questions in single-image and image-pair settings, as well as $1{,}057$ open-ended questions for emotion description. EmoGenie-Assess benchmarks continuous affective and aesthetic prediction using $13{,}175$ mean opinion scores (MOSs) aggregated from human ratings in terms of valence, arousal, dominance, and aesthetic quality. We further develop \textbf{EmoProbe}, which combines emotion question answering with response-aligned, dimension-specific regression heads for continuous score prediction. Experimental results show that EmoProbe achieves superior results on emotion-related QA tasks, while its affective and aesthetic predictions correlate better with human ratings, demonstrating its state-of-the-art capabilities of unified coarse-to-fine affective understanding for AI-generated images.
\end{abstract}

\hypertarget{introduction}{%
\section{Introduction}\label{introduction}}

Recent advances in artificial intelligence-generated content (AIGC) have substantially improved the fidelity, diversity, and controllability of text-to-image (T2I) generation~\citep{Chen2024PixArtSigma,Esser2024SD3,Qwen2025Image}. T2I models can generate high-quality images spanning diverse subjects, scenes, and visual styles from natural-language prompts. The expanding application of these models in visual content creation has increased demand for images that embody specific affective intentions. However, the affect conveyed by a generated image does not always align with the intended emotion, despite semantic fidelity to the prompt. Such discrepancies compromise the intended visual message and undermine the reliability of emotion-conditioned image generation. These concerns extend from emotion generation to emotion understanding. In this context, multimodal large language models (MLLMs) provide a means of interpreting the affective content of generated images, yet their consistency with human perception remains unclear. Accordingly, we consider two fundamental questions:

\emph{Can T2I models generate images that align with the intended emotion?}

\emph{Can MLLMs interpret the affective content of generated images consistently with human perception?}

Together, these questions motivate a unified benchmark that assesses emotion in AI-generated images and supports the evaluation of MLLM emotion understanding.

The affective content of an image can be understood from two distinct perspectives. Expressed emotion denotes the affect conveyed by a depicted individual, while evoked emotion denotes the viewer's affective response to the image. Given their distinct perceptual bases, the two perspectives may yield different judgments for the same image and impose different requirements on emotion generation. Their joint analysis therefore provides a more comprehensive account of affective content. Existing affective datasets predominantly comprise natural images annotated for facial expressions, artistic content, or scene-induced responses~\citep{Mikels2005,Machajdik2010,Peng2015}, whereas research on generated images has focused mainly on expressed emotion in synthetic portraits~\citep{Zhou2024MEMO,Schuhmann2025EmoNet}. Recent MLLM benchmarks incorporate additional task formats for emotion assessment, including ranking, emotion reasoning, dimensional assessment, and open-ended description~\citep{Gao2025EEmoBench,AICA-Bench}. However, a systematic benchmark encompassing both perspectives in AI-generated images remains unavailable. Consequently, the effectiveness of T2I generation and MLLM understanding across expressed and evoked emotions remains insufficiently characterized.

To this end, we construct \textbf{EmoGenie}, a benchmark for jointly evaluating emotion generation and understanding in AI-generated images (AIGIs) from expressed and evoked perspectives. Using $800$ prompts spanning both perspectives, we generate $6{,}400$ images with eight T2I models and collect $384$K human annotations across six applicable dimensions: emotion category, expression intensity, valence, arousal, dominance (VAD), and aesthetic quality. From these annotations, we derive two subsets: \textbf{EmoGenie-QA}, comprising $90$K closed-set questions for single-image and image-pair perception and $1{,}057$ open-ended questions for emotion description; and \textbf{EmoGenie-Assess}, comprising over $13$K continuous affective and aesthetic scores. Together, the subsets support perception, description, and assessment within a unified evaluation framework, enabling joint evaluation of T2I emotion alignment and human-aligned MLLM understanding.

Building on EmoGenie, we further develop \textbf{EmoProbe} through a two-stage training procedure that integrates emotion question answering with continuous score prediction. Specifically, the first stage performs supervised fine-tuning (SFT) on EmoGenie-QA to equip the model with coarse-grained emotion perception capabilities. The second stage uses dimension-specific regression heads to align response-conditioned representations with the perceptual assessments in EmoGenie-Assess, enabling VAD and aesthetic score prediction. This formulation unifies discrete emotion understanding and continuous score prediction within a single multimodal model. Experiments validate the effectiveness of EmoProbe and reveal substantial performance variation across current T2I models and MLLMs.

\begin{table}[t]
\caption{Emotion-analysis datasets and benchmarks. A check indicates an explicitly supported annotation or question format; a cross indicates that the resource does not provide it. QA Pairs reports the number of question--answer pairs. VAD denotes valence, arousal, and dominance; MOS denotes an aesthetic mean opinion score.}
\label{tab:resources}
\centering
\scriptsize
\setlength{\tabcolsep}{3pt}
\resizebox{\linewidth}{!}{%
\begin{tabular}{lccccccccc}
\toprule
\textbf{Resource} & \textbf{Images} & \textbf{QA Pairs} & \textbf{Source} & \textbf{Expressed} & \textbf{Evoked} & \textbf{VAD} & \textbf{MOS} & \textbf{Image Pairs} & \textbf{Open-Ended} \\
\midrule
ArtPhoto~\citep{Machajdik2010} & 807 & $\times$ & Collected & $\times$ & $\checkmark$ & $\times$ & $\times$ & $\times$ & $\times$ \\
Abstract-8~\citep{Machajdik2010} & 280 & $\times$ & Collected & $\times$ & $\checkmark$ & $\times$ & $\times$ & $\times$ & $\times$ \\
Emotion6~\citep{Peng2015} & 1,980 & $\times$ & Collected & $\times$ & $\checkmark$ & $\times$ & $\times$ & $\times$ & $\times$ \\
EEmo-Bench~\citep{Gao2025EEmoBench} & 1,960 & 6,773 & Collected & $\times$ & $\checkmark$ & $\checkmark$ & $\times$ & $\checkmark$ & $\checkmark$ \\
\hdashline
MEMO-Bench~\citep{Zhou2024MEMO} & 7,145 & $\times$ & Generated & $\checkmark$ & $\times$ & $\times$ & $\checkmark$ & $\times$ & $\times$ \\
EmoNet-Face~\citep{Schuhmann2025EmoNet} & 2,500 & $\times$ & Generated & $\checkmark$ & $\times$ & $\times$ & $\times$ & $\times$ & $\times$ \\
\midrule
\textbf{EmoGenie (Ours)} & 6,400 & 91,057 & Generated & $\checkmark$ & $\checkmark$ & $\checkmark$ & $\checkmark$ & $\checkmark$ & $\checkmark$ \\
\bottomrule
\end{tabular}}
\end{table}

Our main contributions are summarized as follows:

\begin{itemize}
    \item We introduce \textbf{EmoGenie}, a large-scale benchmark for emotion analysis in AIGIs, comprising $6{,}400$ images and over $91$K question-answer pairs. By systematically covering both expressed and evoked perspectives, EmoGenie supports emotion perception, description, and assessment within a unified dataset.
    \item We benchmark representative T2I models and MLLMs on EmoGenie, establishing a unified evaluation of emotion generation and understanding. The results reveal marked variation across affective perspectives, emotion categories, and evaluation tasks.
    \item We develop \textbf{EmoProbe}, a multimodal emotion model trained in two stages to integrate emotion question answering with response-aligned regression. EmoProbe achieves state-of-the-art performance on both coarse-grained emotion perception and fine-grained affective assessment of AIGIs.
\end{itemize}

\hypertarget{related-work}{%
\section{Related Work}\label{related-work}}

\subsection{Visual Emotion Datasets}

Traditionally, visual emotion analysis has focused on natural images. Early datasets like ArtPhoto, Abstract-8~\citep{Machajdik2010}, and Emotion6~\citep{Peng2015} capture discrete emotions or affective distributions, while EEmo-Bench~\citep{Gao2025EEmoBench} expands the scope with diverse tasks such as emotion ranking and dimensional assessment. Furthermore, recent benchmarks like AICA-Bench~\citep{AICA-Bench} and E$^3$mo-Bench~\citep{E3mo-Bench} advance the field by distinguishing between expressed emotions and evoked emotions. However, these comprehensive evaluation paradigms remain largely absent for AIGIs. Current AIGI benchmarks rarely combine diverse tasks with joint expressed-evoked modeling, resulting in an incomplete affective characterization.

\subsection{Benchmarks for AIGIs}

Research on affective image generation examines whether synthesized or edited images convey intended emotions. EmoGen~\citep{Yang2024EmoGen} conditions generation on discrete emotion categories, whereas EmotiCrafter~\citep{Dang2025EmotiCrafter} uses continuous valence--arousal coordinates. EmoEdit~\citep{Yang2025EmoEdit} and Make Me Happier~\citep{Lin2025Make} instead modify existing images to elicit a target affect. MEMO-Bench~\citep{Zhou2024MEMO} evaluates emotion generation and interpretation in synthetic portraits through expression, intensity, and visual quality, while EmoNet-Face~\citep{Schuhmann2025EmoNet} provides an expert taxonomy of facial emotions in generated images. Existing evaluations, however, focus largely on generation controllability, affective editing, or facial expression, leaving the joint evaluation of expressed and evoked emotion in AIGIs underexplored.

\begin{figure}[t]
\centering
\includegraphics[width=\linewidth]{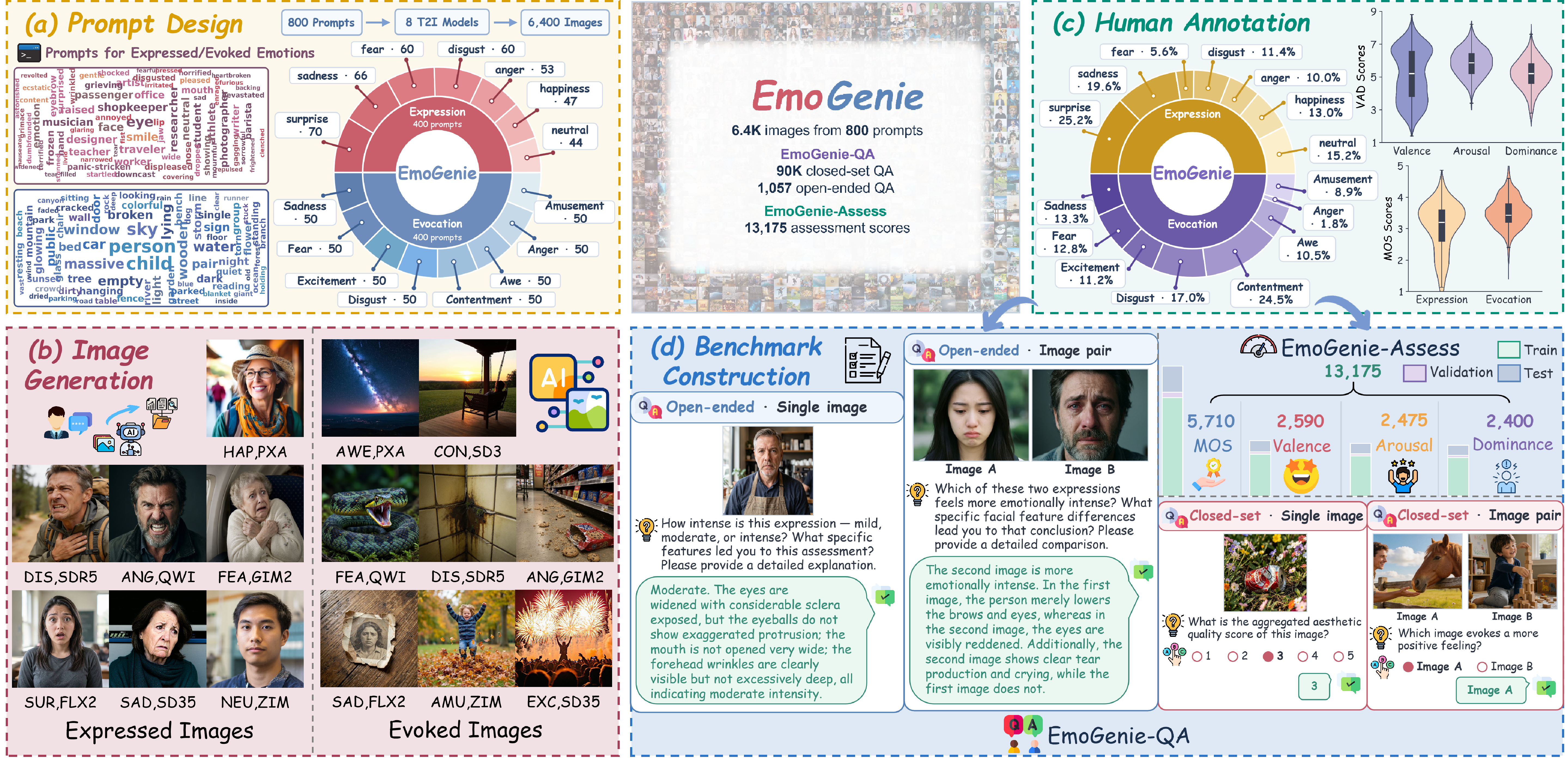}
\caption{EmoGenie construction pipeline. (a) Designs prompts for expressed and evoked emotion. (b) Generates $6{,}400$ images from $800$ prompts with eight T2I models. (c) Collects human annotations of emotion, intensity or VAD, aesthetic quality, and supporting visual evidence. (d) Constructs EmoGenie-QA with $90$K closed-set and $1{,}057$ open-ended QA pairs, and EmoGenie-Assess with $13{,}175$ affective and aesthetic scores.}
\label{fig:dataset_overview}
\end{figure}

\subsection{Emotion-Oriented MLLMs}
Emotion-oriented MLLMs broaden visual emotion analysis beyond recognition by supporting reasoning, explanation, comparison, and continuous assessment. EmoVIT~\citep{Xie2024EmoVIT} and AffectGPT~\citep{Lian2025AffectGPT} are trained with visual instructions designed for emotion understanding, whereas Emotion-Qwen~\citep{Huang2025EmotionQwen} integrates generalist and emotion-specialized experts. EmoCaliber~\citep{Wu2025EmoCaliber} strengthens the reliability of affective judgments through confidence verbalization and calibration. EEmo-Logic~\citep{Gao2026EEmoLogic} adopts multi-stage optimization to progressively learn emotion perception, comparison, description, and continuous affect prediction. Most of these models, however, are developed and evaluated on natural images or address only one affective perspective. Their alignment with human judgments of expressed and evoked emotion in AIGIs therefore remains unclear.

\hypertarget{emoduet-benchmark-construction}{%
\section{EmoGenie Benchmark Construction}\label{sec:emogenie}}

EmoGenie has two subsets: 1) \textbf{EmoGenie-QA} supports emotion perception and evidence-grounded description, whereas 2) \textbf{EmoGenie-Assess} provides continuous affective and aesthetic scores. Fig.~\ref{fig:dataset_overview} summarizes image generation, annotation, and subset construction.

\hypertarget{data-collection}{%
\subsection{Data Collection}\label{data-collection}}

Expressed and evoked emotion rely on different perceptual cues. We therefore construct a separate prompt pool for each perspective. The expressed emotion pool follows Ekman's six basic emotions and neutral~\citep{Ekman1971}: anger (ANG), disgust (DIS), fear (FEA), happiness (HAP), sadness (SAD), surprise (SUR), and neutral (NEU). To represent expressions of varying intensity, we vary affective terms together with the corresponding facial and bodily actions. For evoked emotion, we adopt the discrete taxonomy introduced by ~\citet{Mikels2005}. The positive categories are amusement (AMU), awe (AWE), contentment (CON), and excitement (EXC), and the negative categories are anger, disgust, fear, and sadness. Prompts describe emotionally salient scenes and contextual cues intended to elicit each response. Together, the two pools contain $800$ prompts, with $400$ for each perspective.

We apply all $800$ prompts to eight representative T2I models spanning open- and closed-source systems (Tab.~\ref{tab:t2i_models}). The resulting collection comprises $6{,}400$ AIGIs, evenly distributed across the expressed and evoked perspectives, and supports the systematic evaluation of emotion understanding in MLLMs.  

\begin{table}[t]
\caption{Details of T2I models used for the generation. Year denotes the model release year, and resolution denotes the generation setting used for benchmark construction. Open-source models provide publicly released checkpoints.}
\label{tab:t2i_models}
\centering
\small
\setlength{\tabcolsep}{5pt}
\resizebox{0.8\textwidth}{!}{%
\begin{tabular}{lllcc}
\toprule
\textbf{Type} & \textbf{Label} & \textbf{T2I model} &
\textbf{Year} & \textbf{Resolution} \\
\midrule
Open-source
& PXA  & PixArt-$\Sigma$~\citep{Chen2024PixArtSigma}
& $2024$ & $1024\times1024$ \\
Open-source
& SD3  & Stable Diffusion 3~\citep{Esser2024SD3}
& $2024$ & $1024\times1024$ \\
Open-source
& SD35 & Stable Diffusion 3.5~\citep{StabilityAI2024SD35}
& $2024$ & $1024\times1024$ \\
Open-source
& FLX2 & FLUX.2~\citep{BFL2025Flux2}
& $2025$ & $1360\times1360$ \\
Open-source
& ZIM  & Z-Image~\citep{Tongyi2025ZImage}
& $2025$ & $1024\times1024$ \\
Open-source
& QWI  & Qwen-Image~\citep{Qwen2025Image}
& $2025$ & $1328\times1328$ \\
\midrule
Closed-source
& SDR5 & Seedream 5.0~\citep{ByteDance2026Seedream5}
& $2026$ & $1024\times1024$ \\
Closed-source
& GIM2 & GPT-Image-2~\citep{OpenAI2026GPTImage2}
& $2026$ & $1024\times1024$ \\
\bottomrule
\end{tabular}%
}
\end{table}

\hypertarget{subjective-annotation}{%
\subsection{Subjective Annotation}\label{subjective-annotation}}

Annotation is conducted in two rounds. In the first round, each of the $6{,}400$ images receives $15$ independent annotations. The annotation protocol encompasses three aspects:
    1) \textbf{Emotion category} identifies either the emotion conveyed by the visible expression or the emotion evoked in the viewer, according to the target  affective perspective.
    2) \textbf{Graded affective attributes} comprise expression intensity for expressed-emotion images and VAD for evoked-emotion images. Expression intensity captures the strength of observable facial and bodily cues, whereas VAD characterizes the pleasantness, activation, and perceived control associated with the viewer's affective response using the $9$-point Self-Assessment Manikin scale~\citep{Bradley1994}.
    3) \textbf{Aesthetic quality} is rated on a $1$--$5$ scale according to composition, color, illumination, clarity, realism, and overall visual impact~\citep{ITUBT500}.

The first round therefore yields $96$K image-level annotation records and $384$K individual dimension-level judgments. For each image $i$, let $\ell_i^{(a)}$ denote the raw emotion-category label assigned by annotator $a$, and let $r_{i,d}^{(a)}$ denote the corresponding numerical rating for scalar dimension $d$. We then aggregate the categorical and scalar annotations as
\begin{equation}
\mathcal{C}_i=\underset{c}{\arg\max}\sum_{a=1}^{15}\mathbb{I}\!\left[\ell_i^{(a)}=c\right],
\qquad
\bar{r}_{i,d}=\frac{1}{15}\sum_{a=1}^{15}r_{i,d}^{(a)}.
\end{equation}

The set $\mathcal{C}_i$ contains every category receiving the largest vote count. Images with multiple consensus categories remain in the dataset but are excluded from benchmark questions requiring a unique reference answer. Detailed annotation instructions and agreement analysis are provided in Appendix~\ref{app:annotation_interface}--\ref{app:aggregation_agreement}.

The second round elicits evidence-grounded descriptions for a stratified subset comprising 300 single-image cases and 100 image pairs, corresponding to 500 images in total. The questions follow a structured judgment-to-evidence reasoning design: annotators first identify or compare the relevant emotion or attribute and then justify the judgment using observable visual cues. Their responses describe facial, bodily, compositional, and scene-level evidence; for image pairs, they additionally identify the salient differences between the images. These annotations provide the questions and reference answers for the description task.

\subsection{EmoGenie-QA}

Derived from the aggregated labels and evidence-grounded descriptions, EmoGenie-QA comprises over $91$K question--answer pairs for closed-set perception and open-ended description.

\paragraph{Perception.}
To assess both absolute and relative judgments, we construct single-image and image-pair questions from the aggregated annotations across all applicable dimensions. Single-image questions evaluate whether a model can recognize the affective and aesthetic attributes of an individual image. Image-pair questions instead probe whether it can distinguish variations in these attributes across images. The perception task contains $90$K questions, comprising $40$K single-image and $50$K image-pair questions.

\paragraph{Description.}
To extend evaluation beyond categorical selection, we construct single-image and image-pair questions from the evidence-grounded annotations collected in the second round. Single-image questions require models to identify emotions and explain the relevant visual cues. Image-pair questions require comparative judgments grounded in salient differences between the images. After quality screening, the description task contains $1{,}057$ questions, comprising $957$ single-image and $100$ image-pair questions. All description questions are reserved entirely for evaluation, and their reference answers are excluded from training. Representative examples are shown in Fig.~\ref{fig:dataset_overview}.

\subsection{EmoGenie-Assess}

Whereas EmoGenie-QA evaluates perception and description, EmoGenie-Assess supports fine-grained continuous prediction using the aggregated scalar ratings. It contains approximately $13.2$K image--dimension scores, covering valence, arousal, and dominance for evoked-emotion images and aesthetic quality for both perspectives. Each score is the mean of $15$ independent ratings and retains its original scale: $1$--$9$ for VAD and $1$--$5$ for aesthetic quality. This subset measures how closely model predictions align with the graded affective and aesthetic assessments of human observers.

\hypertarget{emoprobe-two-stage-training-framework}{%
\section{The EmoProbe Approach}\label{sec:emoprobe}}

\begin{figure}[t]
\centering
\includegraphics[width=\linewidth]{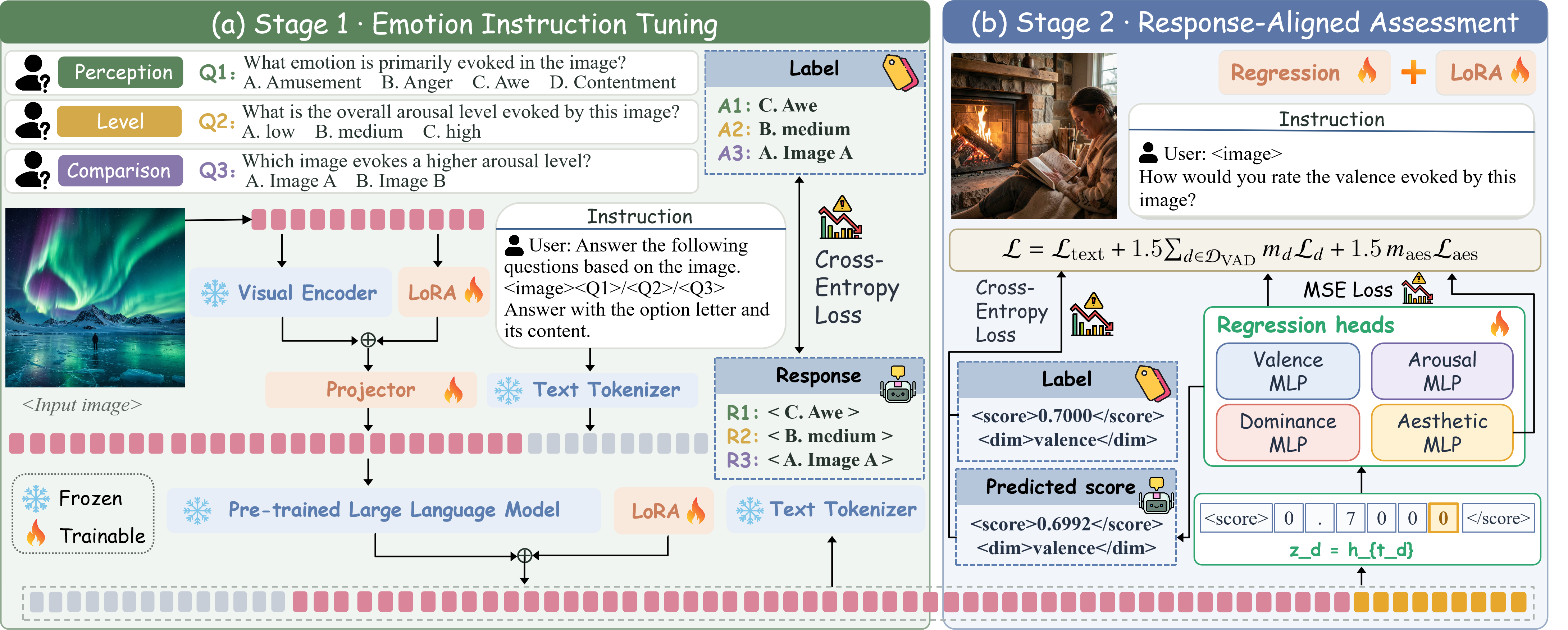}
\caption{Overview of EmoProbe training. Stage $1$ instruction-tunes the vision encoder, multimodal projector, and language model for single-image and image-pair emotion perception. In Stage $2$, the projector is frozen, the vision and language LoRA parameters remain trainable, and four dimension-specific regression heads map response-aligned representations to continuous affective and aesthetic scores. The Stage $2$ objective combines response cross-entropy with masked regression losses.}
\label{fig:emoprobe_framework}
\end{figure}

In this section, we present the architecture and two-stage training of EmoProbe, an emotion-oriented MLLM designed for coarse-grained emotion perception and fine-grained continuous assessment. Fig.~\ref{fig:emoprobe_framework} summarizes the shared multimodal backbone, the two-stage optimization procedure, and the response-aligned regression readout.

\subsection{Model Architecture}
\label{sec:model_architecture}

\paragraph{Overall architecture.}
EmoProbe extends a pretrained MLLM with dimension-specific regression heads, supporting emotion perception, evidence-grounded description, and continuous assessment within a unified model. Given a visual input and task instruction, the multimodal backbone produces contextual representations for textual generation, while response-aligned representations are mapped to continuous affective and aesthetic scores.

\paragraph{Multimodal encoding and fusion.}
EmoProbe adopts InternVL3.5-8B-Instruct~\citep{Wang2025InternVL35} as its multimodal backbone. The model takes an image set $\mathcal{I}$ and an instruction $q$ as input, where $\mathcal{I}$ contains either a single image or an image pair. The vision encoder $E_v$ extracts image features, while the tokenizer $T$ converts the instruction into language tokens. The projector $P$ maps the image features into the embedding space of the language model $F$, which jointly contextualizes the visual and instruction tokens:
\begin{equation}
V=P\!\left(E_v(\mathcal{I})\right),
\qquad
H=F\!\left([V;T(q)]\right),
\end{equation}
where $H$ denotes the resulting sequence of contextual representations. For image-pair inputs, visual tokens from both images are jointly encoded within a shared multimodal context, enabling direct comparison of their affective attributes.

\paragraph{Response-aligned regression.}
To extend textual generation to continuous assessment, EmoProbe incorporates four independently parameterized regression heads for valence, arousal, dominance, and aesthetic quality. For each assessment query, the language model produces a structured response containing a numerical judgment. The hidden state aligned with the final numerical token constitutes the score representation processed by the corresponding regression head. For dimension $d$, let $t_d$ denote the position of this token and $h_{t_d}$ its contextual representation. The associated two-layer multilayer perceptron (MLP) estimates the score as
\begin{equation}
z_d=h_{t_d},
\qquad
\hat{y}_d=W_d^{(2)}\,\mathrm{GELU}\!\left(W_d^{(1)}z_d+b_d^{(1)}\right)+b_d^{(2)}.
\end{equation}

\subsection{Two-Stage Training}
\label{sec:two_stage_training}

EmoProbe follows a coarse-to-fine training strategy. Stage $1$ establishes coarse-grained perception through instruction tuning on EmoGenie-QA, and Stage $2$ extends this capability to fine-grained score regression using EmoGenie-Assess.

\paragraph{Stage $1$: emotion instruction tuning.}
We instruction-tune EmoProbe on the perception portion of EmoGenie-QA. Conditioned on either a single image or an image pair together with a task instruction, the model predicts the corresponding categorical response. The autoregressive cross-entropy loss over the target response tokens jointly optimizes the multimodal projector and the low-rank adaptation (LoRA) parameters introduced into the vision encoder and language model. This optimization adapts the multimodal representations and generated responses to both single-image judgments and image-pair comparisons.

\paragraph{Stage $2$: continuous assessment.}
We subsequently train EmoProbe on EmoGenie-Assess for fine-grained continuous assessment. Given an image and a dimension-specific instruction, the model generates a structured response containing the normalized score. The response-aligned representation defined in Section~\ref{sec:model_architecture} is then mapped by the corresponding regression head to a continuous prediction. The structured response is supervised autoregressively, and the predicted score is regressed against the associated human rating. Throughout this stage, the multimodal projector remains frozen, while the vision and language LoRA parameters and the four regression heads are jointly optimized.
The resulting Stage $2$ objective combines the text and regression losses:
\begin{equation}
\mathcal{L}=\mathcal{L}_{\mathrm{text}}
+\lambda_{\mathrm{VAD}}\!\sum_{d\in\mathcal{D}_{\mathrm{VAD}}}m_d\mathcal{L}_d
+\lambda_{\mathrm{aes}}m_{\mathrm{aes}}\mathcal{L}_{\mathrm{aes}},
\label{eq:stage2_objective}
\end{equation}
where $\mathcal{D}_{\mathrm{VAD}}$ comprises valence, arousal, and dominance; $\mathcal{L}_{\mathrm{text}}$ denotes the autoregressive cross-entropy loss; and $\mathcal{L}_d$ denotes the mean-squared error for dimension $d$. The masks $m_d$ and $m_{\mathrm{aes}}$ retain only regression terms with available targets. VAD supervision is therefore restricted to evoked-emotion images, whereas aesthetic supervision applies to both affective perspectives. We set $\lambda_{\mathrm{VAD}}=\lambda_{\mathrm{aes}}=1.5$.

\section{Experiments}
\label{sec:experiments}

\begin{figure*}[t]
    \centering
    \includegraphics[width=\textwidth]{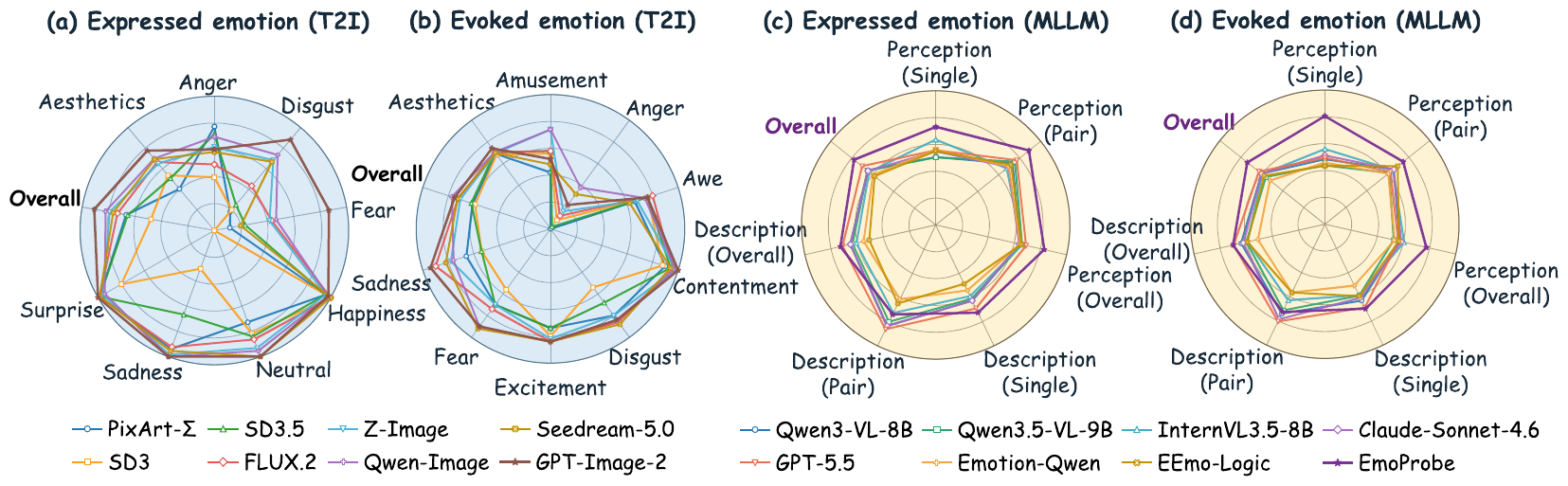}
    \caption{Performance of T2I models and MLLMs from the expressed- and evoked-emotion perspectives. Panels (a) and (b) compare emotion alignment and aesthetic quality across the eight T2I models for expressed and evoked emotion, respectively. Panels (c) and (d) compare MLLM performance for expressed and evoked emotion, respectively, across single-image, image-pair, and overall perception and description scores. The aggregate Overall axis denotes the mean of the perception and description scores. }
    \label{fig:radar_overview}
\end{figure*}

\subsection{Experimental Setup}
\label{sec:experimental_setup}

\paragraph{Implementation details.}
Following Sec.~\ref{sec:two_stage_training}, EmoProbe is trained on two NVIDIA RTX A6000 GPUs with $48$ GB of memory each. Both stages apply LoRA to the linear layers of the vision encoder and language model ($r=64$, $\alpha=128$, dropout $=0.05$). Stage $1$ trains the LoRA parameters and multimodal projector for one epoch with a learning rate of $1\times10^{-4}$, a warm-up ratio of $0.03$, cosine scheduling, weight decay of $0.1$, and an effective batch size of $16$. Stage $2$ freezes the projector and trains the LoRA parameters and four regression heads for one epoch with a learning rate of $5\times10^{-5}$.

\paragraph{Evaluation data and models.}
For emotion generation, we evaluate the eight T2I models using the complete EmoGenie image set. For emotion understanding, we compare EmoProbe with $14$ general-purpose open-source, proprietary, and emotion-oriented MLLMs on EmoGenie-QA and EmoGenie-Assess. The perception and assessment subsets are partitioned into image-disjoint training and test sets at a $4{:}1$ ratio, whereas the description subset is reserved entirely for evaluation.

\paragraph{Evaluation metrics.}
T2I models are evaluated by soft alignment accuracy and mean aesthetic quality. For MLLMs, perception is evaluated by exact-match accuracy, and assessment is evaluated using the Spearman (SRCC) and Pearson (PLCC) correlations. With reference to EEmo-Logic~\citep{Gao2026EEmoLogic}, description is evaluated along four dimensions: \textit{correctness}, \textit{completeness}, \textit{relevance}, and \textit{conciseness}. DeepSeek-v4-Flash~\citep{deepseek-v4} scores the first three, while conciseness is length-based. The four $0$--$2$ scores are averaged and rescaled to $0$--$100$. For open-source MLLMs that cannot directly produce calibrated continuous scores, we adopt the Q-BENCH~\citep{Zhang2024QBenchPlus} logit-softmax protocol. Detailed evaluation procedures are provided in the appendix.

\subsection{Evaluation of T2I Emotion Generation}
\label{sec:t2i_emotion_generation}

\begin{table}[t]
\centering
\caption{Target-emotion alignment and aesthetic quality for eight T2I models on EmoGenie. Alignment is soft accuracy (\%): the target emotion shares the highest human vote count. Avg. pools the $400$ images from each perspective, whereas Aggregate pools all $800$ images. Aesthetics is the mean human rating on a $1$--$5$ scale. Best results are in \textbf{bold}, and second-best results are \underline{underlined}.}
\label{tab:t2i_emotion_generation}
\scriptsize
\setlength{\tabcolsep}{2.2pt}
\renewcommand{\arraystretch}{1.12}
\resizebox{\textwidth}{!}{%
\begin{tabular}{@{}l|cccccccc:c|ccccccccc:c|c:c@{}}
\hline
\textbf{Perspectives}
& \multicolumn{9}{c|}{\textbf{Expressed emotion}}
& \multicolumn{10}{c|}{\textbf{Evoked emotion}}
& \multicolumn{2}{c}{\textbf{Overall}} \\
\cline{1-1}\cline{2-10}\cline{11-20}\cline{21-22}
\textbf{Dimensions}
& \multicolumn{8}{c:}{\textbf{Emotion alignment}}
& \multicolumn{1}{c|}{\textbf{Aesthetics}}
& \multicolumn{9}{c:}{\textbf{Emotion alignment}}
& \multicolumn{1}{c|}{\textbf{Aesthetics}}
& \multicolumn{1}{c:}{\multirow{2}{*}{\shortstack[c]{\textbf{Aggregate}\\[-1pt]\textbf{alignment}}}}
& \textbf{Aesthetics} \\
\cline{1-1}\cline{2-9}\cline{11-19}
\textbf{Models / Metrics}
& ANG & DIS & FEA & HAP & NEU & SAD & SUR & Overall & (1--5)
& AMU & ANG & AWE & CON & DIS & EXC & FEA & SAD & Overall & (1--5)
& & (1--5) \\
\hline
PixArt-$\Sigma$ & \textbf{77.4\%} & 18.3\% & 11.7\% & 95.7\% & 72.7\% & 93.9\% & \underline{98.6\%} & 66.8\% & 2.02 & 42.0\% & 0.0\% & 66.0\% & 90.0\% & 80.0\% & 74.0\% & 70.0\% & 66.0\% & 61.0\% & 3.45 & 63.9\% & 2.74 \\
SD3 & 39.6\% & 20.0\% & 0.0\% & \textbf{100.0\%} & 81.8\% & 30.3\% & 80.0\% & 48.0\% & 2.68 & 56.0\% & 8.0\% & \underline{76.0\%} & 88.0\% & 54.0\% & 80.0\% & 56.0\% & 54.0\% & 59.0\% & 3.42 & 53.5\% & 3.05 \\
SD3.5 & \underline{71.7\%} & 25.0\% & 23.3\% & 95.7\% & 84.1\% & 66.7\% & \underline{98.6\%} & 65.5\% & 2.54 & \underline{58.0\%} & 2.0\% & \underline{76.0\%} & 94.0\% & 68.0\% & 74.0\% & 70.0\% & 54.0\% & 62.0\% & 3.52 & 63.8\% & 3.03 \\
FLUX.2 & 49.1\% & 43.3\% & 43.3\% & \textbf{100.0\%} & 86.4\% & 92.4\% & \underline{98.6\%} & 73.2\% & 3.32 & \underline{58.0\%} & 12.0\% & \textbf{80.0\%} & 96.0\% & \underline{86.0\%} & \textbf{84.0\%} & 74.0\% & \underline{90.0\%} & 72.5\% & 3.52 & 72.9\% & 3.42 \\
Z-Image & 62.3\% & 68.3\% & 41.7\% & 95.7\% & 93.2\% & \underline{98.5\%} & 95.7\% & 79.2\% & 3.25 & \textbf{74.0\%} & 16.0\% & 68.0\% & \underline{98.0\%} & 80.0\% & \underline{82.0\%} & 70.0\% & 78.0\% & 70.8\% & 3.48 & 75.0\% & 3.37 \\
Qwen-Image & 69.8\% & \underline{73.3\%} & \underline{46.7\%} & \underline{97.9\%} & \underline{95.5\%} & \textbf{100.0\%} & 94.3\% & \underline{82.2\%} & \underline{3.49} & \textbf{74.0\%} & \textbf{38.0\%} & 74.0\% & \underline{98.0\%} & 84.0\% & \textbf{84.0\%} & \underline{90.0\%} & 76.0\% & \textbf{77.2\%} & \underline{3.53} & \underline{79.8\%} & \underline{3.51} \\
Seedream-5.0 & 58.5\% & 66.7\% & 20.0\% & \textbf{100.0\%} & \textbf{100.0\%} & 95.5\% & \underline{98.6\%} & 76.5\% & 3.46 & 48.0\% & \underline{32.0\%} & 62.0\% & 94.0\% & \textbf{88.0\%} & \textbf{84.0\%} & \textbf{92.0\%} & 82.0\% & 72.8\% & 3.49 & 74.6\% & 3.48 \\
GPT-Image-2 & 60.4\% & \textbf{88.3\%} & \textbf{86.7\%} & \underline{97.9\%} & \textbf{100.0\%} & \textbf{100.0\%} & \textbf{100.0\%} & \textbf{90.8\%} & \textbf{3.88} & 52.0\% & 22.0\% & \underline{76.0\%} & \textbf{100.0\%} & 84.0\% & \textbf{84.0\%} & \underline{90.0\%} & \textbf{94.0\%} & \underline{75.2\%} & \textbf{3.71} & \textbf{83.0\%} & \textbf{3.80} \\
\hline
\end{tabular}}
\end{table}
\FloatBarrier

Tab.~\ref{tab:t2i_emotion_generation} presents a quantitative comparison of emotion alignment and aesthetic quality across the eight T2I models, while Fig.~\ref{fig:radar_overview}(a) and (b) visualize their performance profiles under the expressed-emotion and evoked-emotion perspectives.

\noindent\textbf{Expressed-emotion alignment.}
For expressed emotion, GPT-Image-2 attains the highest average alignment, followed by Qwen-Image. Seven of the eight models perform better from the expressed perspective than from the evoked perspective, although the magnitude of this advantage differs considerably across models. The category-level results reveal a pronounced disparity in generation difficulty. Happiness and surprise are reproduced consistently across most models, while fear and disgust show much wider variation. Several stronger models also achieve high alignment for neutral and sadness. The reliable generation of sadness alongside unstable fear and disgust indicates that emotional polarity alone is insufficient to explain the observed difficulty; the distinctiveness of facial and bodily configurations appears to play an important role.

\noindent\textbf{Evoked-emotion alignment.}
Evoked-emotion generation presents a different pattern. Qwen-Image obtains the highest average alignment, while GPT-Image-2 ranks second despite its clear advantage for expressed emotion. Contentment is generated reliably by all models, and excitement also remains comparatively stable. Anger is consistently the most difficult category; amusement and sadness exhibit greater model-dependent variation. Evoked emotion is shaped by the joint contribution of scene content, composition, lighting, and atmosphere. Variation across these categories consequently reflects the greater difficulty of controlling the emotion elicited through scene-level visual cues. The change in model ranking further indicates that strong expressed-emotion generation does not consistently transfer to the evoked perspective.

\noindent\textbf{Aesthetic quality.}
Aesthetic ratings reveal a separate source of variation. GPT-Image-2 receives the highest ratings from both perspectives, while Qwen-Image and Seedream-5.0 also produce visually strong results. Seven models receive higher ratings for evoked-emotion images than for expressed-emotion images, with GPT-Image-2 as the exception. This asymmetry is consistent with the additional structural constraints involved in rendering coherent faces, bodies, and emotion-specific details. Together with the perspective-specific alignment patterns, these results point to distinct priorities for further improvement: finer structural control over faces and bodies for expressed emotion, and more reliable affective composition of complete scenes for evoked emotion.

\subsection{Evaluation on MLLM Emotion Understanding}\label{mllm-emotion-understanding}

Fig.~\ref{fig:radar_overview}(c) and (d) depict the corresponding MLLM profiles for expressed and evoked emotion across the single-image, image-pair, and overall perception and description metrics.

% Requires: \usepackage{booktabs}, \usepackage{graphicx},
%           \usepackage[table]{xcolor}, and \usepackage{natbib}

\textbf{Closed-set perception and open-ended description.}
Tab.~\ref{tab:emogenie_qa} reports perception and description results for both expressed and evoked emotion. On closed-set perception, EmoProbe achieves $82.87\%$ for expressed emotion and $77.62\%$ for evoked emotion, outperforming the strongest respective baselines, GPT-5.5 and InternVL3.5-8B-Instruct, by $13.79$ and $17.14$ percentage points. On open-ended description, EmoProbe also ranks first in both perspectives, scoring $72.79\%$ for expressed emotion and $70.08\%$ for evoked emotion, compared with $71.21\%$ and $69.73\%$ for GPT-5.5. Across perception and description, EmoProbe achieves the highest overall score of $75.54\%$. The description results are obtained without description-specific supervision, suggesting that EmoProbe's affective understanding transfers beyond the supervised task formats.

\begin{table*}[t]
\caption{MLLM evaluation on the EmoGenie-QA subset. Perception reports exact-match accuracy (\%), while description reports a normalized evaluation score (\%). Aggregate Overall is the mean of Aggregate Perception and Aggregate Description. Random-guess accuracy is computed by averaging the reciprocal of the number of answer choices across perception questions. }
\label{tab:emogenie_qa}
\centering
\scriptsize
\begingroup
\setlength{\tabcolsep}{3.7pt}
\renewcommand{\arraystretch}{1.08}
\newcommand{\qagroup}[1]{%
  \hdashline
  \multicolumn{16}{@{}l@{}}{\color{black!72}\itshape
  \rule[-0.45ex]{0pt}{2.25ex}#1}\\
  \hdashline}
\newcommand{\qafirstgroup}[1]{%
  \multicolumn{16}{@{}l@{}}{\color{black!72}\itshape
  \rule[-0.45ex]{0pt}{2.25ex}#1}\\
  \hdashline}

\resizebox{\textwidth}{!}{%
\begin{tabular}{@{}l|ccc:ccc|ccc:ccc|c:c:c@{}}
\hline
\textbf{Categories}
& \multicolumn{6}{c|}{\textbf{Expressed emotion}}
& \multicolumn{6}{c|}{\textbf{Evoked emotion}}
& \multicolumn{3}{c}{\textbf{Aggregate}} \\
\cline{1-1}\cline{2-7}\cline{8-13}\cline{14-16}

\textbf{Tasks}
& \multicolumn{3}{c:}{\textbf{Perception}}
& \multicolumn{3}{c|}{\textbf{Description}}
& \multicolumn{3}{c:}{\textbf{Perception}}
& \multicolumn{3}{c|}{\textbf{Description}}
& \multirow{2}{*}{\textbf{Perception}}
& \multirow{2}{*}{\textbf{Description}}
& \multirow{2}{*}{\textbf{Overall}} \\
\cline{1-1}\cline{2-4}\cline{5-7}\cline{8-10}\cline{11-13}

\textbf{Models}
& \textit{Single}$\uparrow$
& \textit{Pair}$\uparrow$
& \textit{Overall}$\uparrow$
& \textit{Single}$\uparrow$
& \textit{Pair}$\uparrow$
& \textit{Overall}$\uparrow$
& \textit{Single}$\uparrow$
& \textit{Pair}$\uparrow$
& \textit{Overall}$\uparrow$
& \textit{Single}$\uparrow$
& \textit{Pair}$\uparrow$
& \textit{Overall}$\uparrow$
& & & \\
\hline

\textit{Random guess}
& 35.98\% & 50.00\% & 44.71\%
& -- & -- & --
& 38.37\% & 50.00\% & 44.20\%
& -- & -- & --
& 44.43\% & -- & -- \\
\hline

\qafirstgroup{General-purpose open-source MLLMs}

Janus-Pro-7B~\citep{Chen2025JanusPro}
& 44.95\% & 59.68\% & 54.12\%
& 55.77\% & 68.75\% & 57.39\%
& 39.94\% & 53.72\% & 46.85\%
& 58.73\% & 56.80\% & 58.59\%
& 50.09\% & 58.13\% & 54.11\% \\

LLaVA-OneVision-1.5-8B-Instruct~\citep{An2025LLaVAOneVision15}
& 52.81\% & 58.60\% & 56.42\%
& 58.68\% & 70.45\% & 60.14\%
& 44.39\% & 57.38\% & 50.90\%
& 62.02\% & 60.10\% & 61.87\%
& 53.36\% & 61.22\% & 57.29\% \\

Qwen2.5-VL-7B-Instruct~\citep{Bai2025Qwen25VL}
& 57.03\% & 71.82\% & 66.24\%
& 50.46\% & 66.45\% & 52.46\%
& 47.43\% & 64.02\% & 55.75\%
& 56.40\% & 62.20\% & 56.84\%
& 60.43\% & 55.18\% & 57.80\% \\

Qwen3-VL-8B-Instruct~\citep{Bai2025Qwen3VL}
& 50.53\% & 73.20\% & 64.65\%
& 62.73\% & \underline{83.65\%} & 65.34\%
& 48.47\% & 65.58\% & 57.05\%
& 63.13\% & 74.05\% & 63.96\%
& 60.44\% & 64.48\% & 62.46\% \\

Qwen3.5-VL-9B~\citep{Qwen2026Qwen35}
& 50.33\% & 75.50\% & 66.01\%
& 61.29\% & 80.15\% & 63.64\%
& 43.85\% & 61.36\% & 52.63\%
& 59.04\% & 71.20\% & 59.97\%
& 58.59\% & 61.36\% & 59.98\% \\

InternVL3-8B-Instruct~\citep{Zhu2025InternVL3}
& 61.49\% & 64.08\% & 63.10\%
& 58.26\% & 66.20\% & 59.25\%
& 51.25\% & 62.18\% & 56.73\%
& 59.93\% & 61.25\% & 60.03\%
& 59.57\% & 59.73\% & 59.65\% \\

InternVL3.5-8B-Instruct~\citep{Wang2025InternVL35}
& \underline{63.54\%} & 67.56\% & 66.04\%
& 58.93\% & 72.90\% & 60.67\%
& 56.03\% & 64.90\% & \underline{60.48\%}
& 58.77\% & 62.90\% & 59.08\%
& \underline{62.96\%} & 59.69\% & 61.32\% \\

\qagroup{Proprietary MLLMs}

Claude-Sonnet-4.6~\citep{Anthropic2026Sonnet46}
& 54.76\% & 69.34\% & 63.84\%
& 62.62\% & 83.60\% & 65.24\%
& 51.33\% & 65.54\% & 58.45\%
& 60.90\% & \underline{78.65\%} & 62.25\%
& 60.86\% & 63.38\% & 62.12\% \\

GPT-5.5~\citep{OpenAI2026GPT55}
& 55.61\% & \underline{77.24\%} & \underline{69.08\%}
& \underline{69.05\%} & \textbf{86.35\%} & \underline{71.21\%}
& 49.56\% & 62.54\% & 56.07\%
& \underline{68.84\%} & \textbf{80.45\%} & \underline{69.73\%}
& 61.87\% & \underline{70.29\%} & \underline{66.08\%} \\

\qagroup{Emotion-oriented MLLMs}

Emotion-Qwen~\citep{Huang2025EmotionQwen}
& 56.11\% & 69.64\% & 64.54\%
& 54.20\% & 61.50\% & 55.11\%
& 46.00\% & 60.50\% & 53.27\%
& 50.76\% & 56.50\% & 51.20\%
& 58.29\% & 52.68\% & 55.49\% \\

EmoCaliber~\citep{Wu2025EmoCaliber}
& 62.55\% & 72.88\% & 68.98\%
& 47.64\% & 55.30\% & 48.60\%
& 44.79\% & 66.14\% & 55.50\%
& 55.92\% & 53.65\% & 55.75\%
& 61.51\% & 53.04\% & 57.27\% \\

EEmo-Logic~\citep{Gao2026EEmoLogic}
& 55.12\% & 71.80\% & 65.51\%
& 48.77\% & 65.05\% & 50.80\%
& 43.66\% & \underline{69.48\%} & 56.61\%
& 59.39\% & 56.55\% & 59.17\%
& 60.58\% & 56.00\% & 58.29\% \\

EmoVIT~\citep{Xie2024EmoVIT}
& 52.91\% & 47.72\% & 49.68\%
& 43.74\% & 50.40\% & 44.57\%
& 56.21\% & 48.84\% & 52.52\%
& 46.47\% & 41.10\% & 46.06\%
& 51.25\% & 45.50\% & 48.37\% \\

AffectGPT~\citep{Lian2025AffectGPT}
& 36.96\% & 50.06\% & 45.12\%
& 32.14\% & 45.05\% & 33.75\%
& \underline{56.72\%} & 49.98\% & 53.34\%
& 32.53\% & 49.60\% & 33.83\%
& 49.67\% & 33.80\% & 41.74\% \\

\hdashline
\textbf{EmoProbe (Ours)}
& \textbf{73.02\%} & \textbf{88.84\%} & \textbf{82.87\%}
& \textbf{72.61\%} & 74.05\% & \textbf{72.79\%}
& \textbf{80.55\%} & \textbf{74.70\%} & \textbf{77.62\%}
& \textbf{69.88\%} & 72.55\% & \textbf{70.08\%}
& \textbf{79.96\%} & \textbf{71.11\%} & \textbf{75.54\%} \\
\hline
\end{tabular}}
\endgroup
\end{table*}

\textbf{Complementary single-image and pairwise evaluation.}
The different chance levels of single-image and pairwise questions should be considered when comparing their accuracies in Tab.~\ref{tab:emogenie_qa}. Relative to the corresponding random-guess baselines, EmoProbe achieves similar margins for expressed-emotion perception in the single-image and pairwise settings ($37.04$ vs.\ $38.84$ percentage points), whereas the margins differ substantially for evoked-emotion perception ($42.18$ vs.\ $24.70$ percentage points). Model rankings also change across formats. For example, on evoked-emotion perception, EmoVIT outperforms EEmo-Logic on single-image questions ($56.21\%$ vs.\ $43.66\%$), whereas EEmo-Logic performs better on pairwise questions ($69.48\%$ vs.\ $48.84\%$). These patterns support evaluating both formats, as single-image recognition and cross-image comparison reveal different aspects of emotion understanding.

\begin{table*}[t]
\caption{MLLM evaluation on the EmoGenie-Assess subset. Results report SRCC and PLCC correlations with human ratings; Overall is the unweighted mean across the four dimensions. }
\label{tab:emogenie_assess}
\centering
\scriptsize
\begingroup
\setlength{\tabcolsep}{5.5pt}
\renewcommand{\arraystretch}{1.18}
\newcommand{\assessgroup}[1]{%
  \hdashline
  \multicolumn{11}{@{}l@{}}{\color{black!72}\itshape\rule[-0.45ex]{0pt}{2.25ex}#1}\\
  \hdashline}
\newcommand{\assessfirstgroup}[1]{%
  \multicolumn{11}{@{}l@{}}{\color{black!72}\itshape\rule[-0.45ex]{0pt}{2.25ex}#1}\\
  \hdashline}
\resizebox{\textwidth}{!}{%
\begin{tabular}{@{}l|cc:cc:cc:cc|cc@{}}
\hline
\textbf{Dimensions}
& \multicolumn{2}{c:}{\textbf{Valence}}
& \multicolumn{2}{c:}{\textbf{Arousal}}
& \multicolumn{2}{c:}{\textbf{Dominance}}
& \multicolumn{2}{c|}{\textbf{Aesthetics}}
& \multicolumn{2}{c}{\textbf{Overall}} \\
\cline{1-1}\cline{2-3}\cline{4-5}\cline{6-7}\cline{8-9}\cline{10-11}
\textbf{Models / Metrics}
& \textit{SRCC}$\uparrow$ & \textit{PLCC}$\uparrow$
& \textit{SRCC}$\uparrow$ & \textit{PLCC}$\uparrow$
& \textit{SRCC}$\uparrow$ & \textit{PLCC}$\uparrow$
& \textit{SRCC}$\uparrow$ & \textit{PLCC}$\uparrow$
& \textit{SRCC}$\uparrow$ & \textit{PLCC}$\uparrow$ \\
\hline

Janus-Pro-7B~\citep{Chen2025JanusPro} & 0.727 & 0.577 & 0.380 & 0.294 & 0.089 & 0.085 & $-0.072$ & $-0.044$ & 0.281 & 0.228 \\
LLaVA-OneVision-1.5-8B-Instruct~\citep{An2025LLaVAOneVision15} & 0.783 & 0.760 & 0.394 & 0.393 & $-0.083$ & $-0.061$ & 0.326 & 0.318 & 0.355 & 0.353 \\
Qwen2.5-VL-7B-Instruct~\citep{Bai2025Qwen25VL} & 0.646 & 0.625 & 0.276 & 0.284 & 0.203 & 0.163 & 0.173 & 0.152 & 0.325 & 0.306 \\
Qwen3-VL-8B-Instruct~\citep{Bai2025Qwen3VL} & 0.675 & 0.732 & 0.346 & 0.324 & $-0.172$ & $-0.130$ & 0.317 & \underline{0.337} & 0.291 & 0.316 \\
Qwen3.5-VL-9B~\citep{Qwen2026Qwen35} & 0.758 & 0.761 & 0.232 & 0.255 & 0.075 & 0.039 & 0.169 & 0.202 & 0.308 & 0.314 \\
InternVL3-8B-Instruct~\citep{Zhu2025InternVL3} & 0.733 & 0.774 & 0.373 & 0.384 & 0.028 & 0.007 & 0.102 & 0.087 & 0.309 & 0.313 \\
InternVL3.5-8B-Instruct~\citep{Wang2025InternVL35} & 0.707 & 0.693 & 0.295 & 0.326 & 0.012 & 0.016 & 0.143 & 0.059 & 0.289 & 0.274 \\
EEmo-Logic~\citep{Gao2026EEmoLogic} & \underline{0.790} & \underline{0.789} & \underline{0.429} & \underline{0.426} & \underline{0.513} & \underline{0.503} & \underline{0.355} & 0.315 & \underline{0.522} & \underline{0.508} \\
\textbf{EmoProbe (Ours)} & \textbf{0.836} & \textbf{0.855} & \textbf{0.481} & \textbf{0.499} & \textbf{0.628} & \textbf{0.616} & \textbf{0.637} & \textbf{0.756} & \textbf{0.645} & \textbf{0.681} \\
\hline
\end{tabular}}
\endgroup
\end{table*}

\textbf{Continuous score prediction.}
Tab.~\ref{tab:emogenie_assess} shows that dominance is the weakest VAD dimension for general-purpose MLLMs, with SRCC ranging from $-0.172$ to $0.203$; aesthetic assessment also exhibits low correlations with human ratings. In contrast, the affect-adapted EEmo-Logic and EmoProbe achieve stronger dominance prediction, while arousal is their weakest VAD dimension. For EmoProbe, arousal reaches an SRCC/PLCC of $0.481/0.499$, below its correlations on the other three assessment dimensions, indicating that fine-grained estimation of emotional activation remains challenging. Overall, EmoProbe achieves the strongest agreement with human ratings across all four assessment dimensions, with mean SRCC and PLCC values of $0.645$ and $0.681$, respectively.

% \FloatBarrier
\subsection{Ablation Study}\label{ablation-study}

\begin{table*}[t]
\caption{EmoProbe ablations on EmoGenie. Perception reports exact-match accuracy (\%), description reports a normalized evaluation score (\%), and assessment reports SRCC / PLCC. Variants without regression heads are evaluated on assessment using Q-BENCH, whereas two-stage variants use the learned regression heads. }
\label{tab:ablations}
\centering
\scriptsize
\setlength{\tabcolsep}{3.0pt}
\renewcommand{\arraystretch}{1.10}
\resizebox{\textwidth}{!}{%
\begin{tabular}{@{}c|l|cccc|ccc|ccc|ccccc@{}}
\hline
\multirow{2}{*}{\textbf{No.}}
& \multirow{2}{*}{\textbf{Backbone}}
& \multicolumn{4}{c|}{\textbf{Training components}}
& \multicolumn{3}{c|}{\textbf{Perception}}
& \multicolumn{3}{c|}{\textbf{Description}}
& \multicolumn{5}{c}{\textbf{Assessment}} \\
\cline{3-6}\cline{7-9}\cline{10-12}\cline{13-17}
&
& \textbf{Vision} & \textbf{Projector} & \textbf{LLM} & \textbf{Reg.}
& \textit{Single}$\uparrow$ & \textit{Pair}$\uparrow$ & \textit{Overall}$\uparrow$
& \textit{Single}$\uparrow$ & \textit{Pair}$\uparrow$ & \textit{Overall}$\uparrow$
& \textit{Valence}$\uparrow$ & \textit{Arousal}$\uparrow$ & \textit{Dominance}$\uparrow$
& \textit{Aesthetics}$\uparrow$ & \textit{Overall}$\uparrow$ \\
\hline
(1) & InternVL3.5-8B
& $\times$ & $\times$ & $\checkmark$ & $\times$
& 71.80\% & 81.10\% & 76.97\%
& 68.03\% & 72.05\% & 68.41\%
& 0.746/0.738 & 0.417/0.437 & 0.319/0.377 & 0.261/0.295 & 0.436/0.462 \\
(2) & InternVL3.5-8B
& $\checkmark$ & $\checkmark$ & $\checkmark$ & $\times$
& \underline{77.08\%} & \textbf{81.84\%} & \underline{79.72\%}
& \underline{68.69\%} & \underline{72.60\%} & \underline{69.06\%}
& 0.729/0.677 & 0.404/0.430 & 0.142/0.260 & 0.350/0.331 & 0.406/0.424 \\
\rowcolor[gray]{0.92}
(3) & \textbf{InternVL3.5-8B}
& $\checkmark$ & $\checkmark$ & $\checkmark$ & $\checkmark$
& \textbf{77.70\%} & \underline{81.77\%} & \textbf{79.96\%}
& \textbf{70.88\%} & \textbf{73.30\%} & \textbf{71.11\%}
& \textbf{0.836}/\underline{0.855} & \textbf{0.481}/\textbf{0.499}
& \textbf{0.628}/\textbf{0.616} & \textbf{0.637}/\textbf{0.756}
& \textbf{0.645}/\textbf{0.681} \\
\hdashline
(4) & Qwen3.5-VL-9B
& $\checkmark$ & $\checkmark$ & $\checkmark$ & $\times$
& 69.90\% & 81.13\% & 76.14\%
& 38.12\% & 41.73\% & 38.47\%
& 0.790/0.833 & 0.250/0.312 & 0.348/0.337 & 0.558/0.637 & 0.486/0.530 \\
(5) & Qwen3.5-VL-9B
& $\checkmark$ & $\checkmark$ & $\checkmark$ & $\checkmark$
& 71.30\% & 81.49\% & 76.96\%
& 44.82\% & 56.30\% & 45.91\%
& \underline{0.835}/\textbf{0.862} & \underline{0.451}/\underline{0.466}
& \underline{0.623}/\underline{0.609} & \underline{0.625}/\underline{0.748}
& \underline{0.634}/\underline{0.671} \\
\hline
\end{tabular}}
\end{table*}

% \FloatBarrier

To assess the effectiveness of the principal components in EmoProbe, we conduct the ablation studies summarized in Tab.~\ref{tab:ablations}. Comparing rows $(1)$ and $(2)$ shows that jointly adapting the vision encoder and projector with the LLM improves perception and description, establishing the contribution of multimodal adaptation. Rows $(2)$ and $(3)$ isolate the effect of Stage $2$: incorporating response-aligned regression improves all four assessment dimensions while maintaining perception and further enhancing description. Rows $(4)$ and $(5)$ extend this analysis to Qwen3.5-VL, where the gains obtained after Stage $2$ indicate that response-aligned regression remains effective under a different multimodal backbone. Finally, rows $(3)$ and $(5)$ compare the performance of the two MLLM backbones, with InternVL3.5-8B achieving stronger overall results.

% \hypertarget{conclusion}{%
\section{Conclusion}\label{conclusion}
In this paper, we introduce \textbf{EmoGenie}, a benchmark for evaluating emotion generation and understanding in AIGIs from expressed- and evoked-emotion perspectives. Building on this benchmark, we develop \textbf{EmoProbe}, an emotion-oriented MLLM that integrates coarse-grained perception and fine-grained continuous assessment through two-stage training. Experiments demonstrate the effectiveness of EmoProbe and reveal substantial variation in the emotion-generation capabilities of current T2I models across affective perspectives and emotion categories, providing valuable insights into key challenges and directions for future affect-aware image generation and understanding.

\subsection*{AI use statement}

Generative AI tools were used to support literature retrieval and discovery and to assist with manuscript preparation through grammatical correction and rhetorical refinement. AI assistance was also used to transform human annotations into selected structured question--answer instances and to score model responses in the open-ended evaluation. All AI-assisted benchmark instances and evaluation outputs were reviewed against the underlying human annotations. Generative AI was not used to determine the research questions, experimental design, technical contributions, interpretation of the results, or scientific conclusions. The authors reviewed all AI-assisted work and take full responsibility for the final manuscript, benchmark, analyses, and conclusions.

\subsection*{Ethics statement}

This study focuses on emotion generation by T2I models and emotion understanding by MLLMs, with human participants involved only in the annotation of AI-generated images. Annotators assessed expressed and evoked emotions, expression intensity, VAD, aesthetic quality, and evidence-grounded emotion descriptions. Before data collection, all annotators were informed of the annotation procedure and the intended use of the resulting data; each participant provided informed consent and received monetary compensation. Annotation records were processed without directly identifying personal information. The construction and intended research use of EmoGenie follow the principles of privacy protection, responsible data use, and transparency set out in the ICLR Code of Ethics.

\subsection*{Reproducibility statement}

We have made efforts to ensure the reproducibility of this work. Sec.~\ref{sec:emogenie} describes the construction of EmoGenie, including prompt design, image generation, human annotation, data aggregation, and the organization of EmoGenie-QA and EmoGenie-Assess. Sec.~\ref{sec:emoprobe} specifies the EmoProbe architecture and two-stage training procedure, while Sec.~\ref{sec:experiments} reports the comparison models, evaluation settings, and metrics. The appendix provides additional generation and annotation details, benchmark statistics, training objectives, evaluation criteria, and qualitative examples. Together, the main paper and appendix document the complete pipeline from benchmark construction to model training and evaluation. We plan to release the benchmark, code, prompts, annotations, and evaluation resources upon publication to facilitate further research.

\bibliography{iclr2027_conference}
\bibliographystyle{iclr2027_conference}

\clearpage
\appendix
\section{Detailed EmoGenie Design and Image Generation}
\subsection{Emotion Taxonomies and Prompt Design}
\label{app:prompt_design}

EmoGenie treats expressed emotion and evoked emotion as two distinct objectives during prompt construction because they depend on different sources of visual evidence. Expressed emotion concerns the affect displayed by a depicted person and is primarily conveyed through facial configuration, gaze, posture, and bodily action. Evoked emotion concerns the affect elicited in a viewer and may arise from objects, events, spatial composition, illumination, colour, atmosphere, or their interaction. Accordingly, we construct an independent prompt pool for each perspective.

Across both pools, prompt design follows three principles. First, the intended emotion is stated explicitly to establish a controlled generation objective. Second, each prompt contains concrete visual content that can realize this objective, avoiding prompts that consist only of an abstract emotion word. Third, prompts remain concise enough to leave substantial freedom in composition and rendering. The resulting prompts specify a common affective target without prescribing an identical scene. We supply the same semantic prompts to every T2I model so that their outputs can be compared under matched generation conditions.

\paragraph{Expressed-emotion prompts.}
The expressed-emotion pool follows Ekman's six basic emotions and adds a neutral category~\citep{Ekman1971}, yielding anger, disgust, fear, happiness, sadness, surprise, and neutral. Each prompt combines five elements: an age group, a gender term, a social or occupational identity, the intended expression, and observable evidence supporting that expression. The evidence is written in visual terms so that intensity is reflected in observable facial and bodily actions.

\textit{Mild anger: ``a teenage woman passenger with an annoyed expression, jaw slightly tightened.''}\\
\textit{Intense anger: ``a young woman artist with an enraged expression, eyes glaring, fists clenched.''}

These examples vary the tension of the face, direction of gaze, and involvement of the body while preserving the same emotion category.

The prompt pool spans mild, moderate, and intense expressions. Intensity is represented by coordinated changes in descriptive vocabulary and visible behaviour, including the tension of the mouth or jaw, eye opening and gaze, eyebrow position, tears, hand gestures, and posture. Neutral prompts retain the same structured format but use cues such as relaxed features, a steady gaze, or a calm and still posture. This design makes intensity a controlled prompt attribute while avoiding a single fixed action template for each emotion.

The $400$ expressed-emotion prompts comprise $53$ anger, $60$ disgust, $60$ fear, $47$ happiness, $44$ neutral, $66$ sadness, and $70$ surprise instances. They include $156$ mild, $110$ moderate, and $134$ intense expressions. Demographic and contextual attributes are also varied: the pool contains $193$ prompts referring to women and $207$ referring to men, with $87$ teenage, $110$ young, $106$ middle-aged, and $97$ elderly subjects. Fourteen identities, including passenger, artist, researcher, teacher, athlete, musician, and office worker, diversify the depicted context. These attributes are distributed across emotion and intensity categories instead of being assigned to a single fixed combination.

\paragraph{Evoked-emotion prompts.}
The evoked-emotion pool adopts the eight-category taxonomy of Mikels et al.~\citep{Mikels2005}: amusement, awe, contentment, excitement, anger, disgust, fear, and sadness. It follows a scene-centred design in which semantic content and visual organization jointly establish the intended viewer response. The prompts cover interpersonal events, natural environments, everyday objects, signs of damage or neglect, potentially threatening situations, and atmospheric scenes. Human figures are included when they contribute to the depicted event. Many prompts rely on objects, environments, lighting, or composition as their primary affective evidence.

Prompts within each category cover several elicitation routes. Awe is represented through scale and visual vastness, and contentment through warmth and stillness.

\textit{Awe: ``a massive glacier with deep blue crevasses under a clear sky.''}\\
\textit{Contentment: ``a warm blanket draped over a chair beside a glowing fireplace.''}

Anger prompts depict violations, obstruction, or unfair conduct. Fear prompts introduce threat, uncertainty, darkness, height, or precarious spatial relations.

\textit{Anger: ``a freshly painted wall with graffiti sprayed across it.''}\\
\textit{Fear: ``a long dark corridor with no visible end.''}

Sadness is represented through loss, abandonment, deterioration, and isolation. Amusement draws on playful incongruity and light-hearted interaction, while excitement is associated with anticipation, speed, crowds, performance, and visually dynamic events.

The evoked-emotion pool contains $50$ prompts for each of the eight categories and $400$ prompts in total.

\subsection{T2I Models and Image Generation}
\label{app:t2i_generation}

We select eight T2I models spanning open-source checkpoints and proprietary APIs, several generations of diffusion and rectified-flow systems, and architectures with different efficiency targets. The selection includes established models and more recent systems with distinct architectural and access characteristics. Each model receives the same $800$ semantic prompts and generates one image per prompt, producing $800$ images per model and $6400$ images overall. Open-source models are run with their released checkpoints and native inference pipelines, and proprietary models are accessed through APIs. No candidate reranking or post-generation filtering is performed using emotion labels or aesthetic scores.

\textbf{PixArt-$\Sigma$}~\citep{Chen2024PixArtSigma} is a diffusion-transformer model developed for efficient high-resolution text-to-image synthesis. It adopts a weak-to-strong training strategy that transfers knowledge from lower-resolution and lower-cost training stages to more demanding generation settings. The model supports multi-scale image synthesis and extends the PixArt family towards direct $4$K generation. Its design emphasizes reduced training cost while maintaining prompt adherence and fine visual detail.

\textbf{Stable Diffusion 3}~\citep{Esser2024SD3} introduces the Multimodal Diffusion Transformer architecture and trains it with rectified flow. Text and image representations retain modality-specific parameters while interacting through joint attention, enabling close alignment between linguistic instructions and generated content. The model combines multiple text encoders to capture token-level detail and broader prompt semantics. These components target improved typography, spatial composition, prompt adherence, and visual fidelity.

\textbf{Stable Diffusion 3.5}~\citep{StabilityAI2024SD35} extends the Stable Diffusion 3 family through revised training and model configurations. It retains the multimodal diffusion-transformer formulation while improving prompt adherence, image quality, and stylistic diversity. The family provides models with different capacity and efficiency profiles, including a high-capacity Large variant. Its training places particular emphasis on varied visual styles and more reliable rendering of complex prompts.

\textbf{FLUX.2}~\citep{BFL2025Flux2} belongs to the FLUX family of rectified-flow image generators. It is designed for detailed visual synthesis, instruction following, and high-resolution output across varied subjects and styles. The FLUX.2-klein-$9$B variant emphasizes efficient inference and can generate images with a small number of sampling steps. This efficiency-oriented design distinguishes it from models that rely on longer iterative denoising schedules.

\textbf{Z-Image}~\citep{Tongyi2025ZImage} is an image-generation foundation model based on a single-stream diffusion transformer. Text conditions and noisy visual representations are processed within a unified sequence, allowing cross-modal interaction throughout generation. The model is designed to combine efficient inference with semantic alignment and detailed synthesis. Its architecture provides a compact alternative to diffusion transformers that maintain separate processing streams for different modalities.

\textbf{Qwen-Image}~\citep{Qwen2025Image} is a text-to-image foundation model developed for general image synthesis and precise visual control. It combines a multimodal diffusion-transformer architecture with representations derived from the Qwen vision-language family. The model places particular emphasis on rendering complex text, preserving fine-grained semantics, and following detailed natural-language instructions. It also supports a unified range of generation and image-editing tasks.

\textbf{Seedream 5.0}~\citep{ByteDance2026Seedream5} is a proprietary image-generation model developed for high-fidelity synthesis and detailed instruction following. It supports photographic, artistic, and design-oriented content while preserving fine visual details under complex prompts. The model also provides image-editing capabilities within the same generation system. Its internal architecture and training configuration are not publicly exposed.

\textbf{GPT-Image-2}~\citep{OpenAI2026GPTImage2} is a proprietary model integrated into OpenAI's multimodal image-generation system. It is designed to follow detailed natural-language instructions and preserve relationships among objects, attributes, and spatial constraints. The model supports image synthesis and editing, with particular attention to text rendering and iterative refinement. Its internal architecture, training data, and sampling procedure are not publicly disclosed.

\subsection{Visual-Quality Variation}
\label{app:visual_quality_variation}

EmoGenie retains one generated output for every prompt--model combination without filtering by aesthetic quality. This preserves the quality variation produced under the benchmark generation settings and prevents post-generation selection from altering the empirical model distributions.

\begin{figure*}[t]
    \centering
    \includegraphics[width=0.8\linewidth]{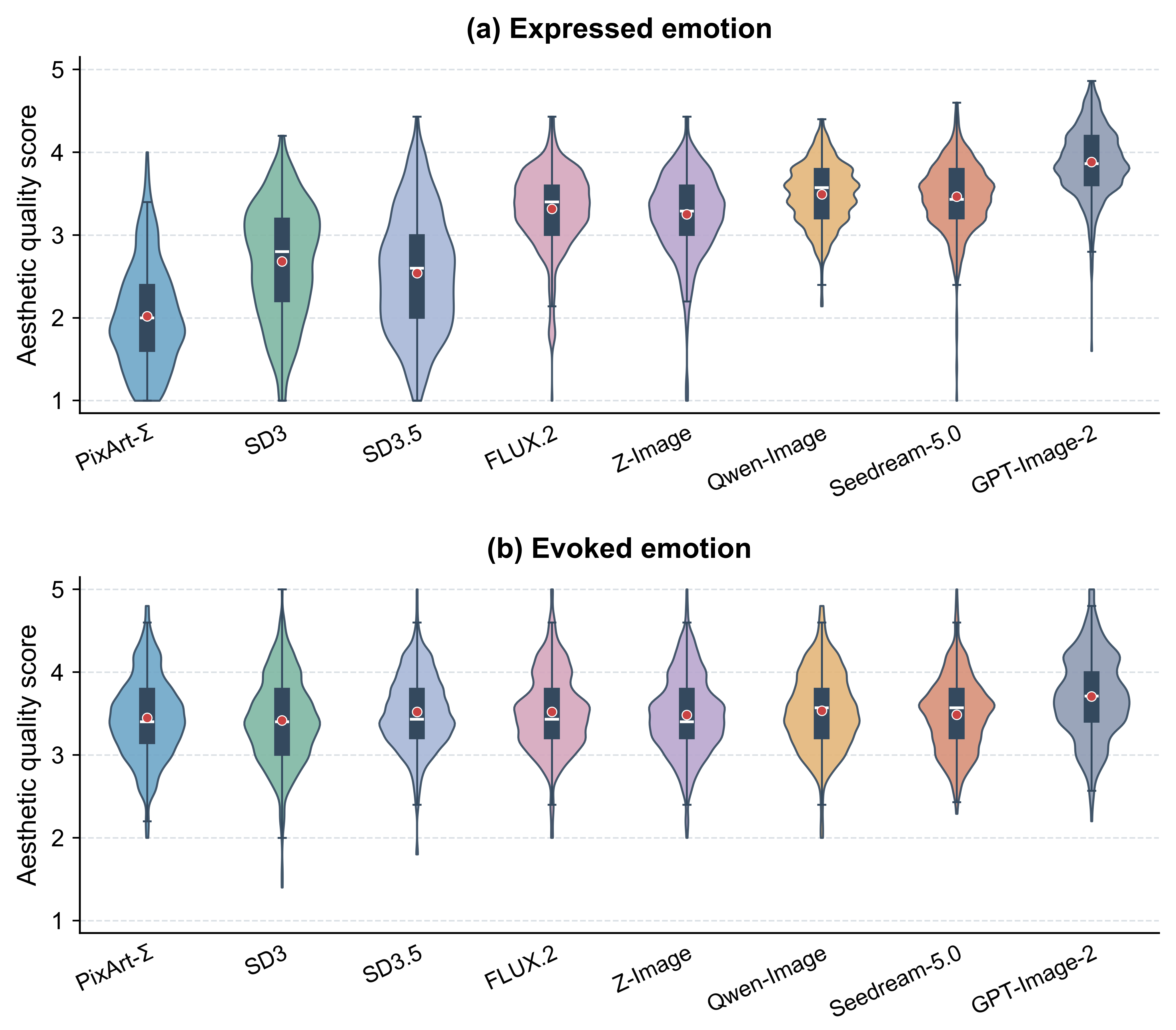}
    \caption{Distribution of aesthetic quality across T2I models for (a) expressed-emotion and (b) evoked-emotion images. Each violin summarizes the human-aggregated ratings of $400$ images from one model. Internal boxes indicate the interquartile range, white lines denote the median, and red points denote the mean. All ratings use the original $1$--$5$ scale.}
    \label{fig:a3_visual_quality}
\end{figure*}

Figure~\ref{fig:a3_visual_quality} shows the image-level distributions of human-aggregated aesthetic ratings for the $400$ expressed-emotion and $400$ evoked-emotion images generated by each model. The expressed-emotion distributions differ substantially across models. GPT-Image-2 attains the highest mean rating ($3.88$), followed by Qwen-Image ($3.49$) and Seedream 5.0 ($3.46$), whereas PixArt-$\Sigma$ has a markedly lower centre ($2.02$). SD3 and SD3.5 also exhibit broad distributions with pronounced lower-quality tails. By contrast, Qwen-Image, Seedream 5.0, FLUX.2, and Z-Image concentrate more of their probability mass above the midpoint of the scale. These distributional differences complement the aggregate results in the main text by showing that the performance gap is not driven by a small number of isolated outputs.

The evoked-emotion distributions are more tightly clustered. Their model means range from $3.42$ to $3.71$, compared with $2.02$ to $3.88$ for expressed emotion, and the interquartile ranges overlap extensively. GPT-Image-2 remains the strongest model on average, but its separation from the remaining models is smaller than in the expressed-emotion setting. Seven models receive higher mean ratings for evoked-emotion images; GPT-Image-2 is the only model with a higher mean for expressed emotion. The largest perspective shifts occur for PixArt-$\Sigma$, SD3.5, and SD3. This pattern is consistent with the additional rendering constraints imposed by expressed-emotion prompts, for which facial structure, bodily coherence, and category-specific expression cues must be jointly preserved. Evoked-emotion prompts can distribute affective evidence across scene content, lighting, colour, and composition, yielding more homogeneous visual-quality distributions across generators.

\section{Details of Human Annotation and Benchmark Assembly}

\subsection{Annotation Dimensions and Interface}
\label{app:annotation_interface}

Annotation is conducted in two rounds. Before each round, annotators receive written definitions, scale anchors, and worked examples that distinguish the emotion expressed by a depicted person from the emotion evoked in a viewer. Images are presented in randomized order, and all fields applicable to the assigned affective perspective must be completed before submission.

\begin{figure*}[t]
    \centering
    \includegraphics[width=\textwidth]{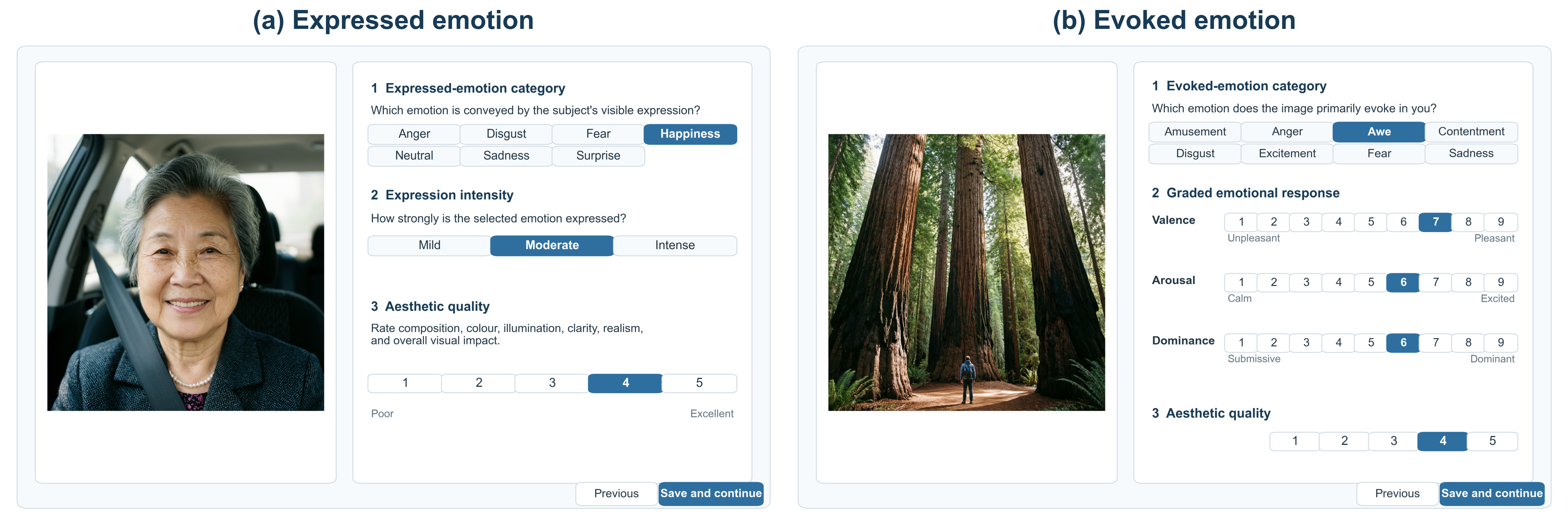}
    \caption{Structured annotation interfaces. (a) Expressed-emotion annotation records the emotion conveyed by the depicted person, its expression intensity, and aesthetic quality. (b) Evoked-emotion annotation records the emotion elicited in the viewer, valence, arousal, dominance, and aesthetic quality. The displayed selections illustrate the interface layout.}
    \label{fig:structured_annotation_interface}
\end{figure*}

The first round collects structured perceptual judgments for all $6400$ images (Fig.~\ref{fig:structured_annotation_interface}). Each image receives $15$ independent annotations, yielding $96$K image-level records and $384$K dimension-level judgments. The interface presents each image alongside the response fields defined for its affective perspective.

For expressed-emotion images, annotators select the emotion conveyed by the subject's visible expression from anger, disgust, fear, happiness, sadness, surprise, and neutral. They then rate expression intensity as mild, moderate, or intense according to the magnitude and coordination of facial and bodily cues. Aesthetic quality is rated on a $1$--$5$ scale based on composition, colour, illumination, clarity, realism, and overall visual impact. Annotators are instructed to assess the emotion displayed by the depicted person rather than their own response to the image.

For evoked-emotion images, annotators select the emotion primarily elicited in them from amusement, anger, awe, contentment, disgust, excitement, fear, and sadness. They also rate valence, arousal, and dominance using the $9$-point Self-Assessment Manikin scale~\citep{Bradley1994}, which measures the pleasantness, activation, and perceived control associated with the elicited response. Aesthetic quality is assessed on the same $1$--$5$ scale. The instructions direct annotators to consider the complete scene, including its content, composition, lighting, colour, and atmosphere.

\begin{figure*}[t]
    \centering
    \includegraphics[width=\textwidth]{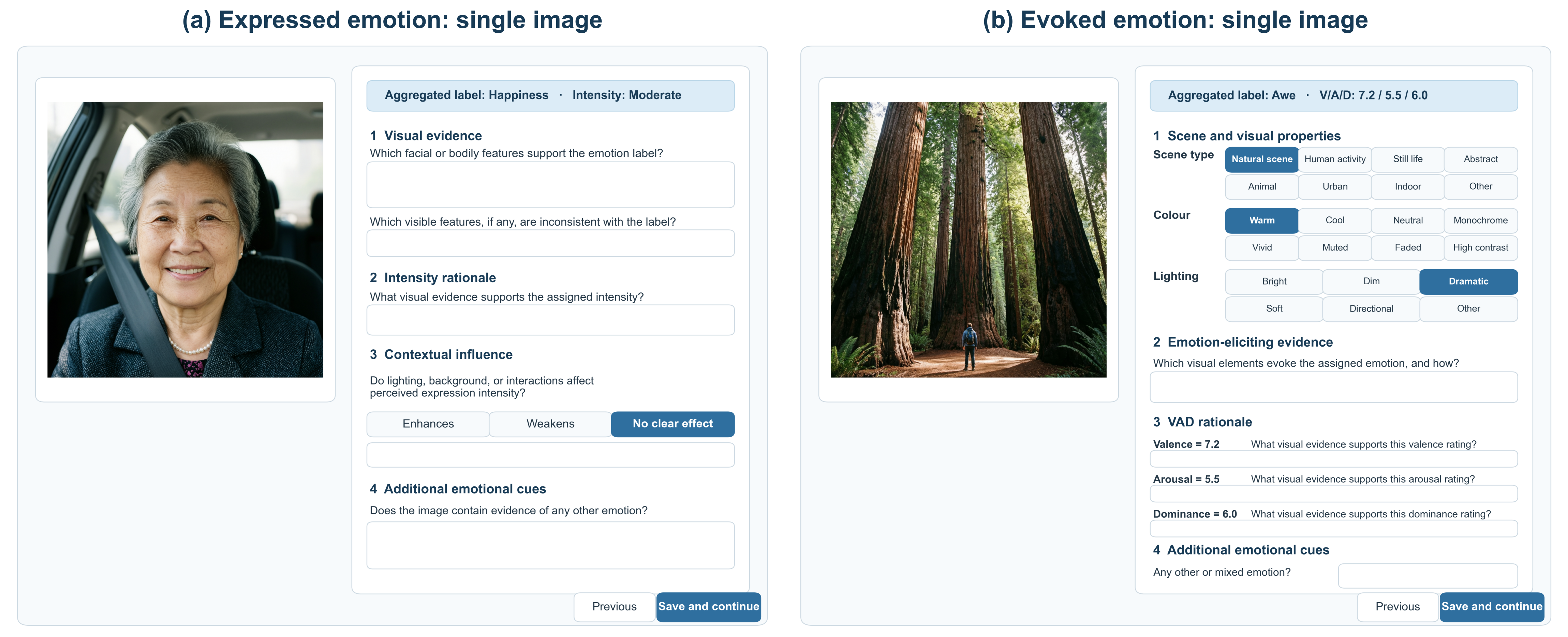}\\[0.5em]
    \includegraphics[width=\textwidth]{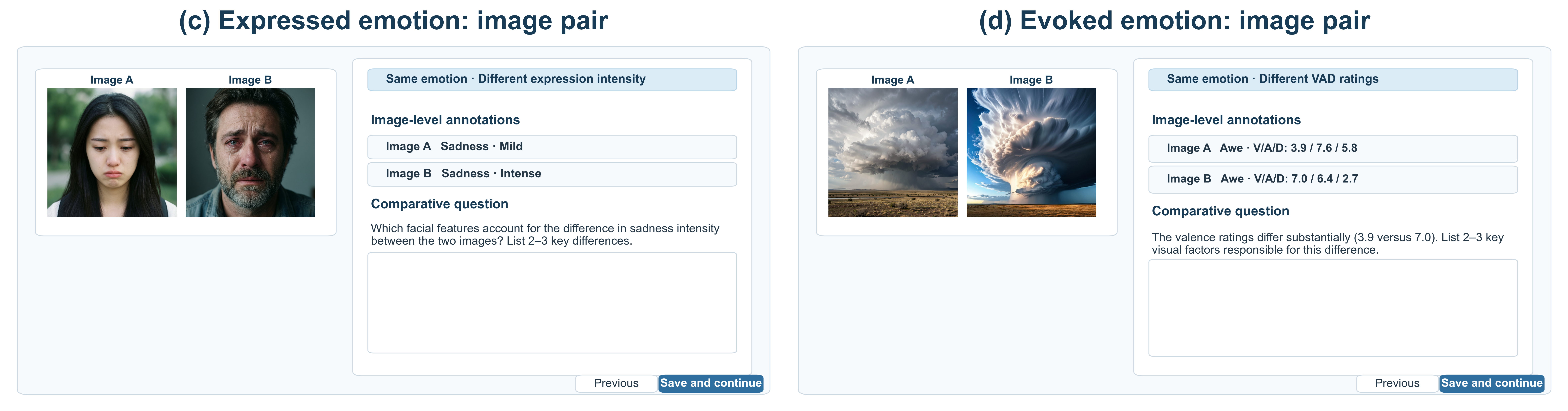}
    \caption{Evidence-grounded description interfaces. (a) Expressed-emotion annotation elicits visual evidence for the category and intensity judgments and records contextual or additional emotional cues. (b) Evoked-emotion annotation characterizes scene-level evidence, elicits rationales for the VAD ratings, and records mixed emotional responses. (c) An expressed-emotion image pair requires a comparison of the facial and bodily evidence underlying an intensity difference. (d) An evoked-emotion image pair requires the visual factors underlying a difference in continuous affective ratings to be identified.}
    \label{fig:open_annotation_interface}
\end{figure*}

The second round elicits evidence-grounded descriptions for a stratified subset of $500$ images, comprising $300$ single-image cases and $100$ image pairs (Fig.~\ref{fig:open_annotation_interface}). Four annotators complete this round. For expressed emotion, the interface provides the aggregated category and intensity and asks annotators to identify supporting or inconsistent facial and bodily cues, explain the assigned intensity, describe contextual influences, and note additional emotional cues. For evoked emotion, annotators characterize the scene, colour, and lighting; identify the visual elements associated with the elicited emotion; justify the VAD ratings; and record additional or mixed responses. Pairwise cases display both images side by side and ask annotators to describe the visual differences underlying the comparative judgment. The resulting descriptions are edited into the questions and reference answers for the open-ended task.

\subsection{Aggregation and agreement analysis}
\label{app:aggregation_agreement}

We verify record completeness, image identity, label validity, and rating ranges before aggregation. Emotion categories are aggregated by majority vote, with all tied leading categories retained; tied cases are excluded only from benchmark questions requiring a unique reference answer. Expression intensity, VAD, and aesthetic quality are averaged on their original scales.

Agreement is estimated from the annotations before question construction. We use Gwet's AC1 for nominal emotion categories and its weighted extension, AC2, for the ordered integer ratings~\citep{Gwet2014}. AC2 is calculated with quadratic weights, which assign progressively larger penalties as the distance between two ratings increases. This treatment preserves the ordering of expression intensity ($1$--$3$), VAD ($1$--$9$), and aesthetic quality ($1$--$5$) without treating adjacent and maximally separated disagreements as equivalent.

Let $q$ be the number of response categories, $\pi_k$ the marginal proportion assigned to category $k$, and $p_{kl}^{\mathrm{o}}$ the observed proportion of rater pairs assigned to categories $k$ and $l$. For a weight matrix $W=(w_{kl})$, the observed and chance-expected weighted agreement are
\begin{equation}
P_{\mathrm{o}}^{w}=\sum_{k=1}^{q}\sum_{l=1}^{q}w_{kl}p_{kl}^{\mathrm{o}},
\qquad
P_{\mathrm{e}}^{w}=\frac{\sum_{k=1}^{q}\sum_{l=1}^{q}w_{kl}}{q(q-1)}
\sum_{k=1}^{q}\pi_k(1-\pi_k),
\end{equation}
and the agreement coefficient is
\begin{equation}
\mathrm{AC}=\frac{P_{\mathrm{o}}^{w}-P_{\mathrm{e}}^{w}}{1-P_{\mathrm{e}}^{w}}.
\end{equation}
For nominal categories, $w_{kl}=\mathbb{I}(k=l)$ and the coefficient reduces to AC1. For ordered ratings, we use quadratic weights
\begin{equation}
w_{kl}=1-\left(\frac{k-l}{q-1}\right)^2,
\end{equation}
yielding AC2.

\begin{table}[t]
\caption{Inter-annotator agreement for the principal annotation dimensions. AC1 is used for nominal emotion categories, and quadratic-weighted AC2 is used for ordered integer ratings.}
\label{tab:annotation_agreement}
\centering
\small
\setlength{\tabcolsep}{5.5pt}
\renewcommand{\arraystretch}{1.08}
\begin{tabular}{lllc}
\toprule
\textbf{Perspective} & \textbf{Dimension} & \textbf{Agreement measure} & \textbf{Estimate} \\
\midrule
\multirow{3}{*}{Expressed emotion}
& Emotion category & Gwet's AC1 & $0.682$ \\
& Expression intensity & Quadratic-weighted AC2 & $0.734$ \\
& Aesthetic quality & Quadratic-weighted AC2 & $0.809$ \\
\midrule
\multirow{5}{*}{Evoked emotion}
& Emotion category & Gwet's AC1 & $0.449$ \\
& Valence & Quadratic-weighted AC2 & $0.784$ \\
& Arousal & Quadratic-weighted AC2 & $0.798$ \\
& Dominance & Quadratic-weighted AC2 & $0.763$ \\
& Aesthetic quality & Quadratic-weighted AC2 & $0.878$ \\
\bottomrule
\end{tabular}
\end{table}

Using these coefficients, we evaluate inter-annotator agreement for each annotation dimension under the two affective perspectives. The results are summarized in Tab.~\ref{tab:annotation_agreement}.Expressed-emotion categories show higher agreement than evoked-emotion categories, consistent with the greater subjectivity of viewer-elicited responses. The ordered dimensions exhibit consistently strong weighted agreement. Aesthetic quality attains the highest AC2 for evoked-emotion images, while expression intensity, VAD, and expressed-image aesthetic quality also show substantial concordance. These coefficients quantify agreement while accounting for the distance between rating categories; they do not require exact equality between adjacent ordinal responses.

\subsection{Detailed Construction of EmoGenie-QA}
\label{app:emogenie_qa}

EmoGenie-QA comprises closed-set perception questions and open-ended description questions derived from the aggregated annotations. The perception task comprises single-image questions in multiple-choice and yes-or-no formats, as well as image-pair questions that require a comparison between two images. Both settings cover the applicable dimensions of emotion category, expression intensity, VAD, and aesthetic quality. This portion contains $90{,}000$ questions, including $40{,}000$ single-image and $50{,}000$ image-pair questions. We use an image-disjoint, stratified split of $72{,}000$ questions for training and $18{,}000$ for testing; the test set contains $8{,}000$ single-image and $10{,}000$ image-pair questions. Images occurring in the same pair are assigned jointly to prevent visual overlap between the two splits.

The description task requires an emotion judgment supported by visible evidence. Single-image questions elicit explanations based on facial, bodily, or scene-level cues, whereas image-pair questions require both a comparative judgment and an account of the salient visual differences. The final set contains $1{,}057$ evaluation-only questions, comprising $957$ single-image and $100$ image-pair questions.

\subsection{Detailed Construction of EmoGenie-Assess}
\label{app:emogenie_assess}

EmoGenie-Assess contains continuous affective and aesthetic scores aggregated from the human ratings. Valence, arousal, and dominance are provided for evoked-emotion images on their original $1$--$9$ scales, while aesthetic quality is provided for both affective perspectives on a $1$--$5$ scale. Each score is the mean rating for one image and dimension. 

The subset contains $13{,}175$ scores, comprising $2{,}590$ valence, $2{,}475$ arousal, $2{,}400$ dominance, and $5{,}710$ aesthetic-quality scores. Using the same image-disjoint partitioning principle as EmoGenie-QA, we allocate $10{,}540$ scores to the training side, of which $535$ form the validation set, and reserve $2{,}635$ for testing. The test portion contains $518$ valence, $495$ arousal, $480$ dominance, and $1{,}142$ aesthetic-quality scores.

\section{EmoProbe Loss Functions}
\label{app:emoprobe_objectives}

EmoProbe is optimized progressively with objectives matched to the two training stages. Stage $1$ uses autoregressive language modelling to learn categorical responses on EmoGenie-QA. For a target response $y=(y_1,\ldots,y_N)$ conditioned on visual input $\mathcal{I}$ and instruction $q$, the token-level cross-entropy loss is
\begin{equation}
\mathcal{L}_{\mathrm{text}}
=-\frac{1}{N}\sum_{t=1}^{N}
\log p_{\theta}\!\left(y_t\mid \mathcal{I},q,y_{<t}\right),
\end{equation}
where $N$ denotes the number of supervised response tokens. Minimizing this objective adapts the generated responses to single-image judgments and image-pair comparisons.

Stage $2$ retains the language objective and adds dimension-specific regression supervision on EmoGenie-Assess. Before regression, a rating $y_{i,d}$ for example $i$ and dimension $d$ is mapped from its original interval $[a_d,b_d]$ to $[0,1]$ as $\widetilde{y}_{i,d}=(y_{i,d}-a_d)/(b_d-a_d)$, where $[a_d,b_d]=[1,9]$ for VAD and $[1,5]$ for aesthetic quality. Let $\widehat{y}_{i,d}$ be the normalized prediction and $m_{i,d}\in\{0,1\}$ indicate whether the target is available. The masked mean-squared loss for dimension $d$ is
\begin{equation}
\mathcal{L}_{d}
=\frac{\sum_{i=1}^{B}m_{i,d}
\left(\widehat{y}_{i,d}-\widetilde{y}_{i,d}\right)^2}
{\sum_{i=1}^{B}m_{i,d}},
\end{equation}
where $B$ is the batch size. The mask restricts VAD supervision to evoked-emotion images while retaining aesthetic supervision for both affective perspectives.

The complete Stage $2$ objective combines response generation with the available regression targets:
\begin{equation}
\mathcal{L}_{\mathrm{Stage\,2}}
=\mathcal{L}_{\mathrm{text}}
+\lambda_{\mathrm{VAD}}\!\sum_{d\in\{v,a,r\}}\mathcal{L}_{d}
+\lambda_{\mathrm{aes}}\mathcal{L}_{\mathrm{aes}},
\end{equation}
where $v$, $a$, and $r$ denote valence, arousal, and dominance, respectively. We set $\lambda_{\mathrm{VAD}}=\lambda_{\mathrm{aes}}=1.5$ to balance continuous-score regression with textual response supervision. Predictions are transformed back to their original scales for evaluation.

\section{Detailed Evaluation Settings and Extended Analysis}
\label{app:evaluation_details}

\subsection{Detailed Information of Compared MLLMs}
\label{app:compared_mllms}

The compared MLLMs encompass general-purpose open-source systems, proprietary systems, and architectures developed specifically for emotion understanding. Collectively, these models cover both general multimodal understanding and emotion-specific modelling, ranging from unified vision-language architectures to specialized mechanisms for affective perception and confidence calibration.

\textbf{Janus-Pro-7B}~\citep{Chen2025JanusPro} is a unified multimodal model for visual understanding and image generation. It employs separate visual encoders for the two tasks, allowing each pathway to learn task-specific representations. Both pathways connect to a shared autoregressive language backbone for multimodal reasoning and synthesis. This decoupled visual design separates the representational requirements of perception from those of generation.

\textbf{LLaVA-OneVision-1.5-8B-Instruct}~\citep{An2025LLaVAOneVision15} is an open vision-language model developed within the LLaVA-OneVision framework. It connects an upgraded visual encoder to a language model through a multimodal projector. Its staged training progressively incorporates single-image, multi-image, and video data. These input types are processed within a unified instruction-following interface.

\textbf{Qwen2.5-VL-7B-Instruct}~\citep{Bai2025Qwen25VL} is a general-purpose MLLM for image, document, and video understanding. Its dynamic-resolution mechanism represents visual inputs with a variable number of tokens according to their spatial dimensions. This design preserves fine-grained visual content while adapting computation to the input. Spatial and temporal representations further support visual grounding and video comprehension.

\textbf{Qwen3-VL-8B-Instruct}~\citep{Bai2025Qwen3VL} is an instruction-tuned vision-language model for fine-grained visual grounding and multimodal reasoning. It accepts both single-image and multi-image inputs. Its instruction tuning covers visual recognition, localization, relational understanding, and reasoning over visual evidence. The language backbone produces structured natural-language responses conditioned on the multimodal context.

\textbf{Qwen3.5-VL-9B}~\citep{Qwen2026Qwen35} is a general-purpose model from the Qwen multimodal family. It integrates visual representations with the language and reasoning capabilities of the Qwen3.5 foundation model. The model supports instruction following, multi-image understanding, and visually grounded reasoning. These capabilities are provided through a unified multimodal conversational interface.

\textbf{InternVL3-8B-Instruct}~\citep{Zhu2025InternVL3} is an instruction-tuned MLLM from the InternVL family. It connects a high-capacity vision encoder to a language model through a learned multimodal interface. Native multimodal training aligns visual and linguistic representations before instruction tuning. The model supports single-image, multi-image, and video understanding within a shared architecture.

\textbf{InternVL3.5-8B-Instruct}~\citep{Wang2025InternVL35} is an open-source MLLM trained through native multimodal pre-training followed by instruction tuning. It extends the InternVL architecture with enhanced visual representation learning, multimodal alignment, and reasoning-oriented optimization. The model accepts single-image, multi-image, and video inputs. These modalities are handled through a common conversational interface for visually grounded instruction following.

\textbf{GPT-5.5}~\citep{OpenAI2026GPT55} is a proprietary general-purpose multimodal model accessed through an official interface. It accepts visual and textual inputs and generates instruction-conditioned natural-language responses. Its capabilities encompass broad visual understanding and multimodal reasoning rather than a dedicated emotion-processing architecture. Detailed information about its training data, internal architecture, and model parameters is not publicly available.

\textbf{Claude-Sonnet-4.6}~\citep{Anthropic2026Sonnet46} is a proprietary multimodal model for visual understanding, instruction following, and reasoning. It processes images jointly with textual context to produce natural-language judgments and explanations. The model supports visually grounded analysis across general application domains. Its training data, architectural details, and model parameters are not publicly disclosed.

\textbf{Emotion-Qwen}~\citep{Huang2025EmotionQwen} is an emotion-oriented MLLM that combines fine-grained affective understanding with general vision-language reasoning. Its mixture-of-experts Hybrid Compressor balances general visual information and emotion-specific features through a learned gating network. A Facial Emotion Capture module emphasizes facial cues associated with affective states. Progressive training with emotion-oriented data and multiple LoRA branches limits interference between specialized and general capabilities.

\textbf{EmoCaliber}~\citep{Wu2025EmoCaliber} is an emotion-oriented MLLM that explicitly verbalizes and calibrates predictive confidence. Built on Qwen2.5-VL, it first acquires structured emotion reasoning through supervised fine-tuning. Semantic and probabilistic priors are subsequently introduced to support confidence estimation. A reinforcement-learning stage jointly optimizes emotion prediction and confidence calibration.

\textbf{EEmo-Logic}~\citep{Gao2026EEmoLogic} is an MLLM developed for image-evoked emotion understanding. Its first stage applies emotion instruction tuning to establish categorical perception and affective reasoning. The second stage employs task-specific reinforcement learning for emotion ranking, VAD prediction, and dominant-emotion classification. The same training framework supports evidence-grounded explanations of evoked emotional responses.

\textbf{EmoVIT}~\citep{Xie2024EmoVIT} adapts the InstructBLIP architecture to visual emotion understanding through instruction tuning. Its training corpus contains categorical, conversational, and reasoning instructions for affective image analysis. The instruction-aware Q-Former is optimized to connect visual features with the language model. The visual encoder and language model remain frozen, concentrating adaptation in the Q-Former.

\textbf{AffectGPT}~\citep{Lian2025AffectGPT} is an emotion-oriented MLLM for affective understanding from audio, video, and text. It introduces multimodal pre-fusion modules that integrate modality-specific evidence before language-model processing. Its fusion mechanisms preserve temporal information and emphasize informative features across modalities. Training on MER-Caption supports natural-language emotion descriptions grounded in multimodal evidence.

\subsection{Extended Evaluation of T2I Emotion Generation}
\label{app:t2i_extended_evaluation}

\begin{figure*}[t]
\centering
\includegraphics[width=\textwidth]{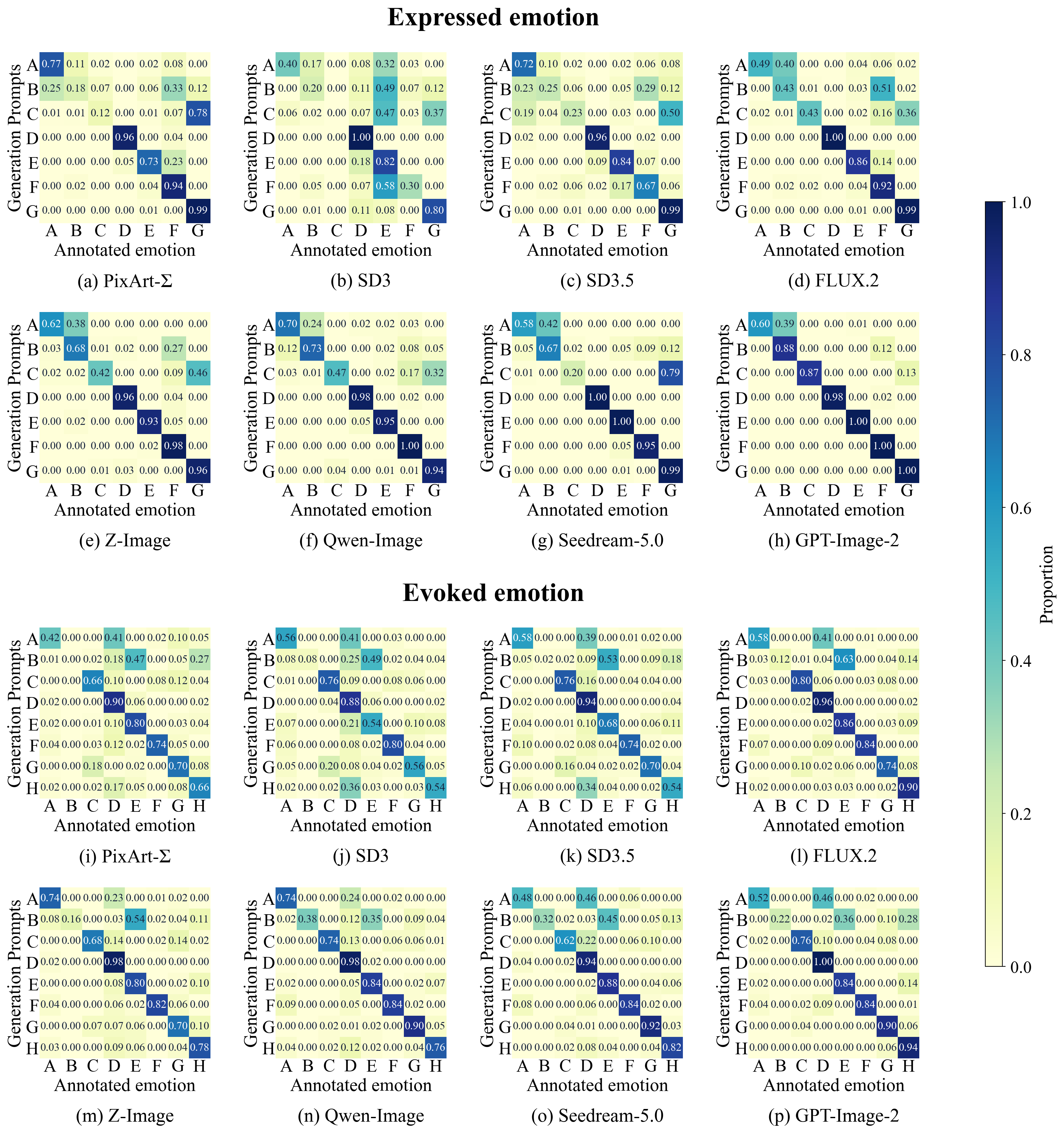}
\caption{Category-level confusion matrices for the emotion-generation capability of eight T2I models. Rows denote generation prompts and columns denote emotions assigned by human annotators. For expressed emotion, A--G correspond to anger, disgust, fear, happiness, neutral, sadness, and surprise. For evoked emotion, A--H correspond to amusement, anger, awe, contentment, disgust, excitement, fear, and sadness. Each row is normalized to sum to $1$; diagonal entries indicate target-emotion alignment, and off-diagonal entries show the distribution of emotion substitutions.}
\label{fig:t2i_emotion_confusion}
\end{figure*}

The main paper reports category-level target-emotion alignment, whereas Fig.~\ref{fig:t2i_emotion_confusion} resolves the human annotations into normalized confusion matrices for each generator. Rows denote the emotions specified by the generation prompts, and columns denote the emotions assigned by annotators. Diagonal concentration therefore indicates reliable target-emotion generation, while off-diagonal mass identifies systematic substitutions between categories.

For expressed emotion, happiness and surprise form the most consistent diagonals across models, whereas the principal errors occur among negative categories with partially shared facial and bodily cues. Anger is frequently annotated as disgust, and fear is often annotated as surprise. GPT-Image-2 separates these categories most clearly, while several other models retain strong alignment for happiness, sadness, and surprise but show concentrated failure on fear or disgust. SD3 exhibits a broader shift towards neutral, suggesting that some generated expressions lack sufficiently distinctive or intense cues.

The evoked-emotion matrices show a different structure. Contentment is generated consistently across models, with awe and excitement also displaying comparatively stable diagonals. Amusement is often interpreted as contentment, indicating that positive scenes may convey pleasantness without the more specific contextual cues associated with amusement. Anger is the least stable category and is frequently reassigned to disgust. The distribution of fear and sadness further differentiates the generators, with stronger models maintaining clearer category boundaries and weaker models shifting these prompts towards awe or contentment.

Overall, expressed-emotion errors cluster among categories with similar facial or bodily manifestations, particularly anger and disgust or fear and surprise. Evoked-emotion errors more often connect states with related valence or scene-level semantics, such as amusement and contentment. These perspective-specific substitutions reveal category-level weaknesses that are obscured by aggregate alignment scores, highlighting the importance of evaluating each affective perspective at the level of individual emotion categories.

\subsection{Evaluation Criteria for Open-Ended Description}
\label{app:description_evaluation}

With reference to the evaluation protocol of EEmo-Logic~\citep{Gao2026EEmoLogic}, we assess open-ended responses along four dimensions: \textit{correctness}, \textit{completeness}, \textit{relevance}, and \textit{conciseness}, each scored from $0$ to $2$. DeepSeek-Flash evaluates the first three dimensions using the question, candidate response, human reference answer, and scoring rubric, without access to the image or model identity. Conciseness is computed deterministically from the candidate-to-reference word-count ratio.

Correctness measures consistency with the reference answer, including emotional judgments, reasoning, and the absence of contradictions or unsupported claims. Completeness measures coverage of the reference answer's key emotional interpretations, reasoning, and VAD information where applicable. Relevance measures how directly the response addresses the question. For these three dimensions, scores of $2$, $1$, and $0$ indicate full satisfaction, partial satisfaction, and substantial failure, respectively. For conciseness, let $r$ denote the candidate-to-reference word-count ratio. The score is $2$ when $2/3 \leq r \leq 2$, $1$ when $1/3 \leq r < 2/3$ or $2 < r \leq 4$, and $0$ otherwise.

Each response is evaluated five times by DeepSeek-Flash. In our implementation, temperature is set to $0$ and top-$p$ to $1$, with reasoning disabled. The first three dimension scores are averaged across the five evaluations. The final response score is the equally weighted mean of all four dimensions, divided by $2$ and multiplied by $100$. Reported results are averaged over the questions in each evaluated subset.

\subsection{Evaluation Criteria for Continuous Assessment}
\label{app:continuous_evaluation_criteria}

We evaluate continuous predictions using Spearman's rank correlation coefficient (SRCC) and Pearson's linear correlation coefficient (PLCC). Let $y_n$ and $\widehat{y}_n$ denote the human-aggregated score and model prediction for sample $n$, respectively, and let $N$ be the number of evaluated samples. SRCC measures monotonic agreement by computing the Pearson correlation between the corresponding ranks $R_n=\operatorname{rank}(y_n)$ and $\widehat{R}_n=\operatorname{rank}(\widehat{y}_n)$:
\begin{equation}
\mathrm{SRCC}=
\frac{\sum_{n=1}^{N}(R_n-\overline{R})(\widehat{R}_n-\overline{\widehat{R}})}
{\sqrt{\sum_{n=1}^{N}(R_n-\overline{R})^2}
 \sqrt{\sum_{n=1}^{N}(\widehat{R}_n-\overline{\widehat{R}})^2}}.
\end{equation}
Average ranks are assigned to tied observations. SRCC approaches $1$ when the predicted ordering closely matches the human-score ordering, even when their numerical relationship is not strictly linear.

PLCC measures linear agreement between the predicted and human scores:
\begin{equation}
\mathrm{PLCC}=
\frac{\sum_{n=1}^{N}(y_n-\overline{y})(\widehat{y}_n-\overline{\widehat{y}})}
{\sqrt{\sum_{n=1}^{N}(y_n-\overline{y})^2}
 \sqrt{\sum_{n=1}^{N}(\widehat{y}_n-\overline{\widehat{y}})^2}},
\end{equation}
where $\overline{y}$ and $\overline{\widehat{y}}$ are the corresponding sample means. A high PLCC indicates that changes in the predicted scores are approximately proportional to changes in the human ratings.

Higher values indicate stronger agreement for both metrics. We compute SRCC and PLCC separately for valence, arousal, dominance, and aesthetic quality. Overall performance is the unweighted mean across the four dimensions, preventing the more numerous aesthetic scores from dominating the aggregate. Both correlations are unchanged by the positive affine transformation between normalized and original scores.

For models without regression heads, we derive continuous predictions using the Q-Bench+ logit-softmax protocol~\citep{Zhang2024QBenchPlus}. Each dimension is represented by a pair of opposite-pole descriptors: \emph{poor}/\emph{good} for valence and aesthetic quality, \emph{calm}/\emph{excited} for arousal, and \emph{submissive}/\emph{dominant} for dominance. Given the next-token logits $\ell_d^{-}$ and $\ell_d^{+}$ for the low- and high-end descriptors of dimension $d$, the probability assigned to the high endpoint is
\begin{equation}
p_d^{+}=\frac{\exp(\ell_d^{+})}
{\exp(\ell_d^{-})+\exp(\ell_d^{+})}.
\end{equation}
We then map this probability to the original interval $[a_d,b_d]$:
\begin{equation}
\widehat{y}_d=a_d+(b_d-a_d)p_d^{+},
\end{equation}
where $[a_d,b_d]=[1,9]$ for VAD and $[1,5]$ for aesthetic quality. This procedure converts the model's relative preference for the two semantic endpoints into a continuous score without training an additional regressor and ensures that larger values consistently denote more positive valence, stronger arousal, greater dominance, or higher aesthetic quality.

\FloatBarrier
\subsection{Dimension-wise Closed-Set QA}\label{app:dimension-wise-closed-qa}

Tabs.~\ref{tab:evoked_dimension_accuracy} and~\ref{tab:expressed_dimension_accuracy} show a consistent pattern: EmoProbe performs strongly on attribute-level and image-pair questions, while single-image emotion identification remains a relative weakness.

For evoked emotion, EmoProbe led 9 of the 10 dimension-by-format comparisons. Its single-image dominance accuracy reached 90.42\%, compared with 65.63\% for the strongest baseline, and its pairwise target-alignment accuracy reached 87.36\%, compared with 80.20\%. The exception was single-image emotion recognition, where EmoCaliber scored 78.91\% against EmoProbe's 75.12\%.

\begin{table*}[t]
\caption{Dimension-wise closed-set QA accuracy for evoked emotion (\%). Single-image and image-pair questions occupy separate column groups. The best result in each column is shown in bold; the bottom row gives the number of questions.}
\label{tab:evoked_dimension_accuracy}
\centering
\scriptsize
\begingroup
\setlength{\tabcolsep}{3pt}
\renewcommand{\arraystretch}{1.08}
\resizebox{\textwidth}{!}{%
\begin{tabular}{@{}lccccc|ccccc@{}}
\toprule
\multirow{2}{*}{\textbf{Model}} & \multicolumn{5}{c|}{\textbf{Single image}} & \multicolumn{5}{c}{\textbf{Image pair}} \\
\cmidrule(lr){2-6}\cmidrule(l){7-11}
 & \shortstack{Evoked\\emotion} & Valence & Arousal & Dominance & Aesthetics & \shortstack{Target\\alignment} & Valence & Arousal & Dominance & Aesthetics \\
\midrule
Janus-Pro-7B & 56.99\% & 37.26\% & 40.51\% & 24.06\% & 42.64\% & 60.81\% & 57.77\% & 52.29\% & 48.18\% & 51.80\% \\
LLaVA-OneVision-1.5-8B-Instruct & 73.10\% & 50.10\% & 32.02\% & 28.33\% & 42.21\% & 56.46\% & 64.26\% & 59.59\% & 51.17\% & 54.93\% \\
Qwen2.5-VL-7B-Instruct & 74.41\% & 51.54\% & 33.43\% & 31.46\% & 49.30\% & 78.79\% & 72.30\% & 62.11\% & 49.39\% & 62.24\% \\
Qwen3-VL-8B-Instruct & 74.05\% & 57.05\% & 44.34\% & 27.60\% & 42.91\% & 78.65\% & 72.67\% & 63.05\% & 56.77\% & 60.91\% \\
Qwen3.5-VL-9B & 77.13\% & 50.48\% & 35.35\% & 23.33\% & 37.83\% & 75.14\% & 72.49\% & 56.22\% & 42.95\% & 64.42\% \\
InternVL3-8B-Instruct & 73.34\% & 53.57\% & 38.28\% & 44.79\% & 49.47\% & 75.98\% & 71.57\% & 63.89\% & 47.15\% & 56.64\% \\
InternVL3.5-8B-Instruct & 71.21\% & 60.91\% & 47.68\% & 38.13\% & 62.70\% & 77.67\% & 72.76\% & 65.67\% & 51.73\% & 60.72\% \\
\midrule
Claude-Sonnet-4.6 & 73.58\% & 58.69\% & 47.68\% & 30.52\% & 48.86\% & 77.25\% & 76.23\% & 61.83\% & 51.45\% & 64.61\% \\
GPT-5.5 & 73.93\% & 57.63\% & 39.19\% & 28.02\% & 51.31\% & 72.89\% & 74.86\% & 61.65\% & 45.94\% & 60.53\% \\
\midrule
Emotion-Qwen & 70.85\% & 51.25\% & 36.06\% & 29.58\% & 45.27\% & 73.31\% & 69.10\% & 58.65\% & 50.14\% & 55.31\% \\
EmoCaliber & \textbf{78.91\%} & 48.94\% & 30.61\% & 22.08\% & 47.20\% & 80.20\% & 74.77\% & 65.01\% & 50.14\% & 65.09\% \\
EEmo-Logic & 70.50\% & 54.83\% & 28.38\% & 24.79\% & 42.82\% & 77.67\% & 74.41\% & 62.96\% & 67.41\% & 67.55\% \\
EmoVIT & 64.34\% & 49.13\% & 42.63\% & 65.21\% & 60.86\% & 51.69\% & 50.09\% & 51.17\% & 50.23\% & 41.84\% \\
AffectGPT & 37.80\% & 52.22\% & 70.51\% & 65.63\% & 55.34\% & 49.86\% & 50.00\% & 49.95\% & 50.05\% & 50.00\% \\
\midrule
\textbf{EmoProbe (Ours)} & 75.12\% & \textbf{80.12\%} & \textbf{85.25\%} & \textbf{90.42\%} & \textbf{72.59\%} & \textbf{87.36\%} & \textbf{78.70\%} & \textbf{68.94\%} & \textbf{70.59\%} & \textbf{72.01\%} \\
\midrule
\emph{No. questions} & 844 & 1,036 & 990 & 960 & 1,142 & 712 & 1,094 & 1,069 & 1,071 & 1,054 \\
\bottomrule
\end{tabular}}
\endgroup
\end{table*}

\begin{table*}[t]
\caption{Dimension-wise closed-set QA accuracy for expressed emotion (\%). Single-image and image-pair questions occupy separate column groups. The best result in each column is shown in bold; the bottom row gives the number of questions.}
\label{tab:expressed_dimension_accuracy}
\centering
\scriptsize
\begingroup
\setlength{\tabcolsep}{3pt}
\renewcommand{\arraystretch}{1.08}
\resizebox{0.75\textwidth}{!}{%
\begin{tabular}{@{}lccc|ccc@{}}
\toprule
\multirow{2}{*}{\textbf{Model}} & \multicolumn{3}{c|}{\textbf{Single image}} & \multicolumn{3}{c}{\textbf{Image pair}} \\
\cmidrule(lr){2-4}\cmidrule(l){5-7}
 & Emotion & \shortstack{Emotion\\intensity} & Aesthetics & \shortstack{Target\\alignment} & \shortstack{Emotion\\intensity} & Aesthetics \\
\midrule
Janus-Pro-7B & 69.44\% & 32.80\% & 32.66\% & 71.06\% & 63.69\% & 49.60\% \\
LLaVA-OneVision-1.5-8B-Instruct & 87.50\% & 36.56\% & 34.68\% & 69.52\% & 64.00\% & 47.45\% \\
Qwen2.5-VL-7B-Instruct & 87.60\% & 35.42\% & 46.67\% & 76.62\% & 79.03\% & 62.31\% \\
Qwen3-VL-8B-Instruct & 86.61\% & 30.30\% & 34.24\% & 81.16\% & 79.35\% & 62.96\% \\
Qwen3.5-VL-9B & 87.10\% & 30.75\% & 32.92\% & 84.17\% & 80.08\% & 66.37\% \\
InternVL3-8B-Instruct & \textbf{90.87\%} & 41.23\% & 51.14\% & 81.16\% & 78.98\% & 40.49\% \\
InternVL3.5-8B-Instruct & 90.08\% & 39.86\% & 58.32\% & 81.44\% & 80.45\% & 47.65\% \\
\midrule
Claude-Sonnet-4.6 & 87.20\% & 33.37\% & 42.56\% & 79.80\% & 80.35\% & 53.10\% \\
GPT-5.5 & 87.40\% & 37.59\% & 41.42\% & 81.98\% & 80.82\% & 71.22\% \\
\midrule
Emotion-Qwen & 83.23\% & 41.46\% & 43.43\% & 74.61\% & 75.88\% & 60.96\% \\
EmoCaliber & 87.60\% & 41.34\% & 56.74\% & 79.25\% & 78.56\% & 63.96\% \\
EEmo-Logic & 84.92\% & 36.56\% & 43.08\% & 77.43\% & 77.30\% & 63.46\% \\
EmoVIT & 52.48\% & 46.47\% & 58.23\% & 49.77\% & 50.24\% & 44.19\% \\
AffectGPT & 26.49\% & 27.68\% & 53.33\% & 50.32\% & 49.97\% & 50.00\% \\
\midrule
\textbf{EmoProbe (Ours)} & 76.09\% & \textbf{72.78\%} & \textbf{70.49\%} & \textbf{89.17\%} & \textbf{85.13\%} & \textbf{92.19\%} \\
\midrule
\emph{No. questions} & 1,008 & 878 & 1,142 & 1,099 & 1,903 & 1,998 \\
\bottomrule
\end{tabular}}
\endgroup
\end{table*}

\FloatBarrier

The expressed-emotion results followed the same pattern: EmoProbe led 5 of 6 comparisons. The clearest gains appeared in single-image intensity (72.78\% versus 46.47\% for EmoVIT) and pairwise aesthetics (92.19\% versus 71.22\% for GPT-5.5). However, InternVL3-8B-Instruct remained stronger on single-image emotion recognition (90.87\% versus 76.09\%). Thus, EmoProbe's aggregate advantage does not imply uniform superiority in categorical recognition.

\FloatBarrier

\section{Qualitative Case Studies}
\label{app:qualitative_results}

\begin{figure*}[t]
\centering
\includegraphics[width=\textwidth]{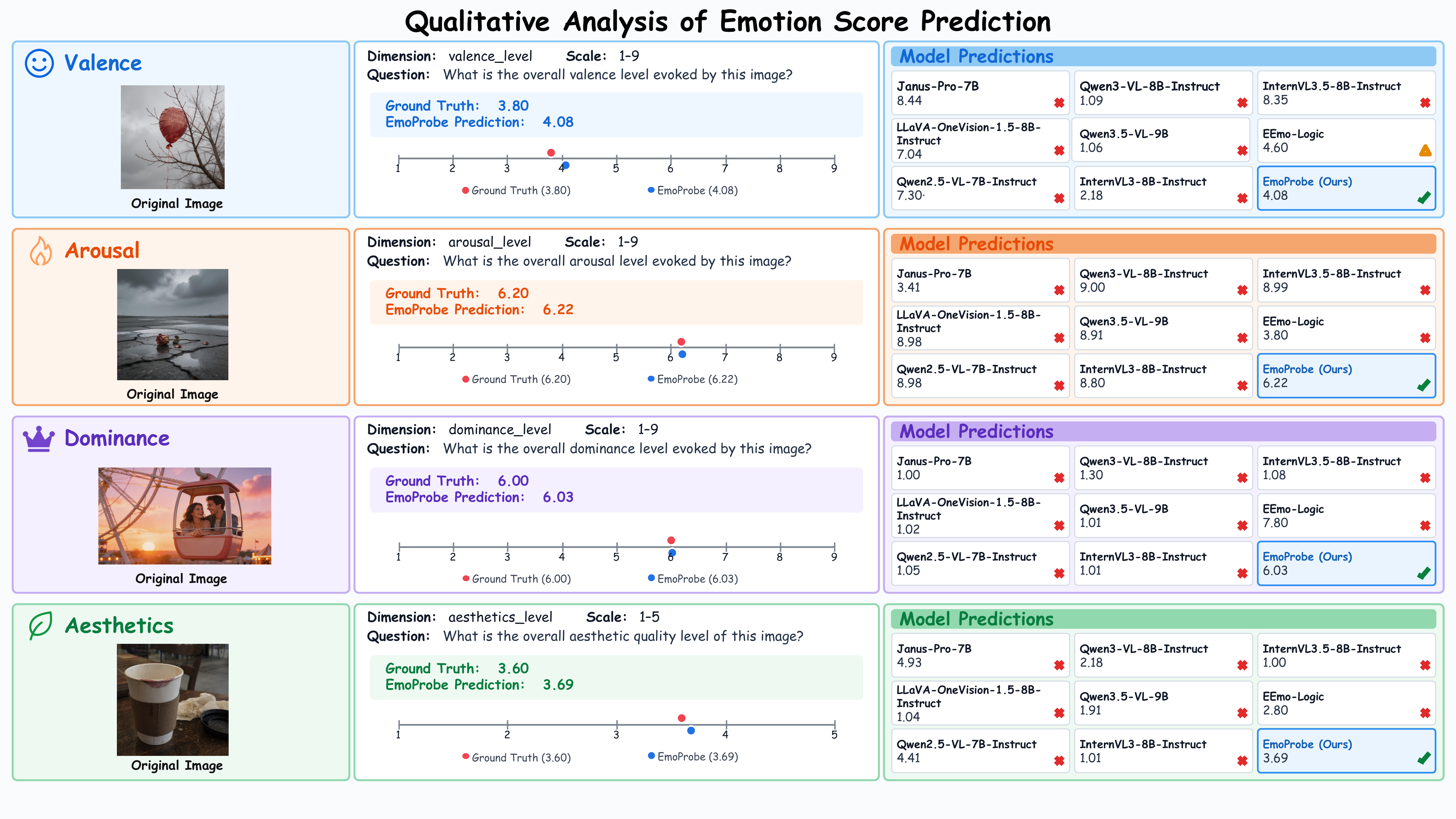}
\caption{Qualitative comparison of continuous affective and aesthetic assessment. Each row presents an example for valence, arousal, dominance, or aesthetic quality and compares the human-aggregated score with predictions from EmoProbe and the evaluated baselines. EmoProbe yields predictions close to the human scores across all four illustrated dimensions.}
\label{fig:qualitative_continuous_assessment}
\end{figure*}

\begin{figure*}[t]
\centering
\includegraphics[width=\textwidth,height=0.88\textheight,keepaspectratio]{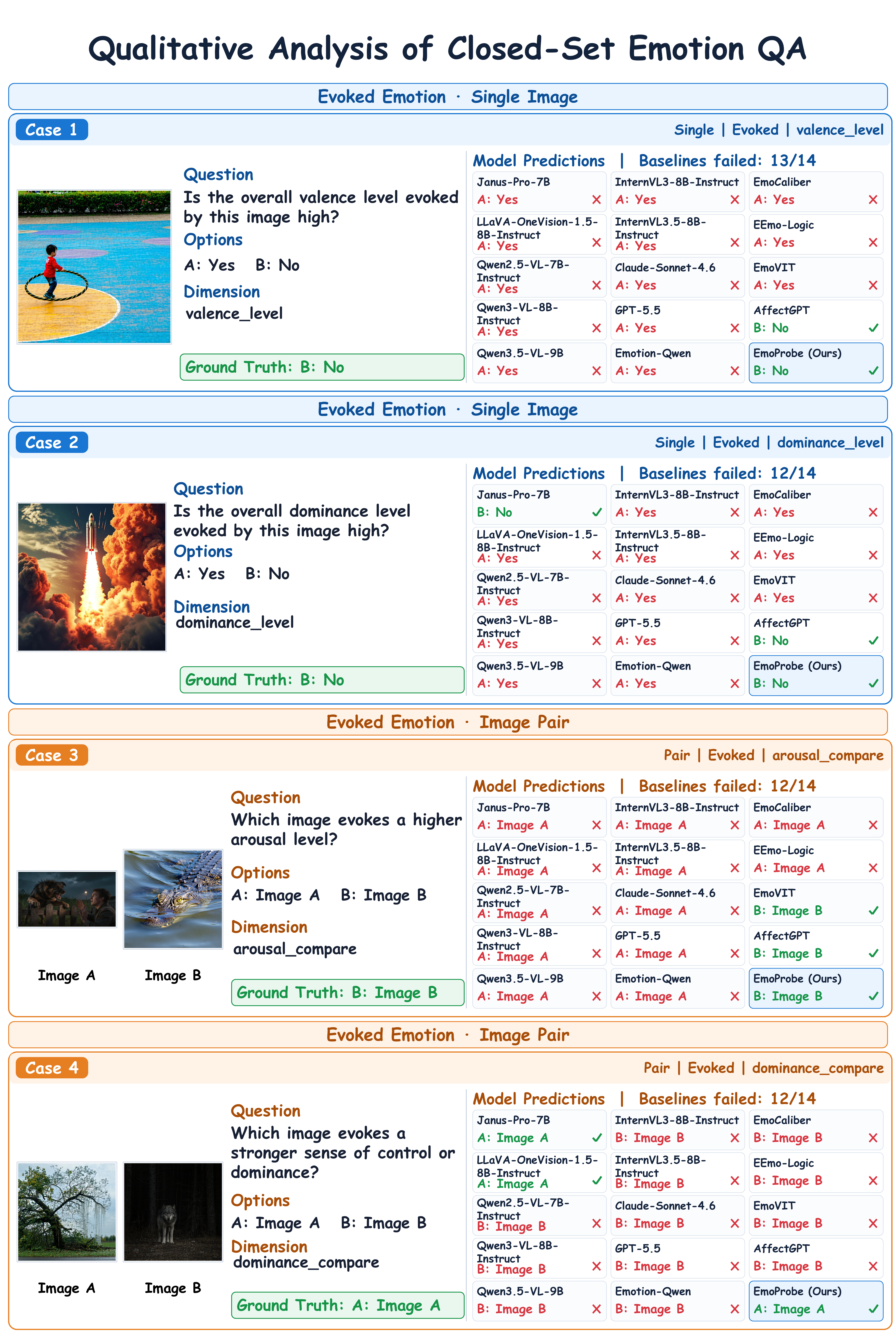}
\caption{Challenging closed-set perception cases for evoked emotion. The examples cover single-image judgments of valence and dominance and image-pair comparisons of arousal and dominance. Green and red markers indicate correct and incorrect responses, respectively; EmoProbe answers all four illustrated questions correctly despite systematic errors among the comparison models.}
\label{fig:qualitative_evoked_qa}
\end{figure*}

\begin{figure*}[t]
\centering
\includegraphics[width=\textwidth,height=0.88\textheight,keepaspectratio]{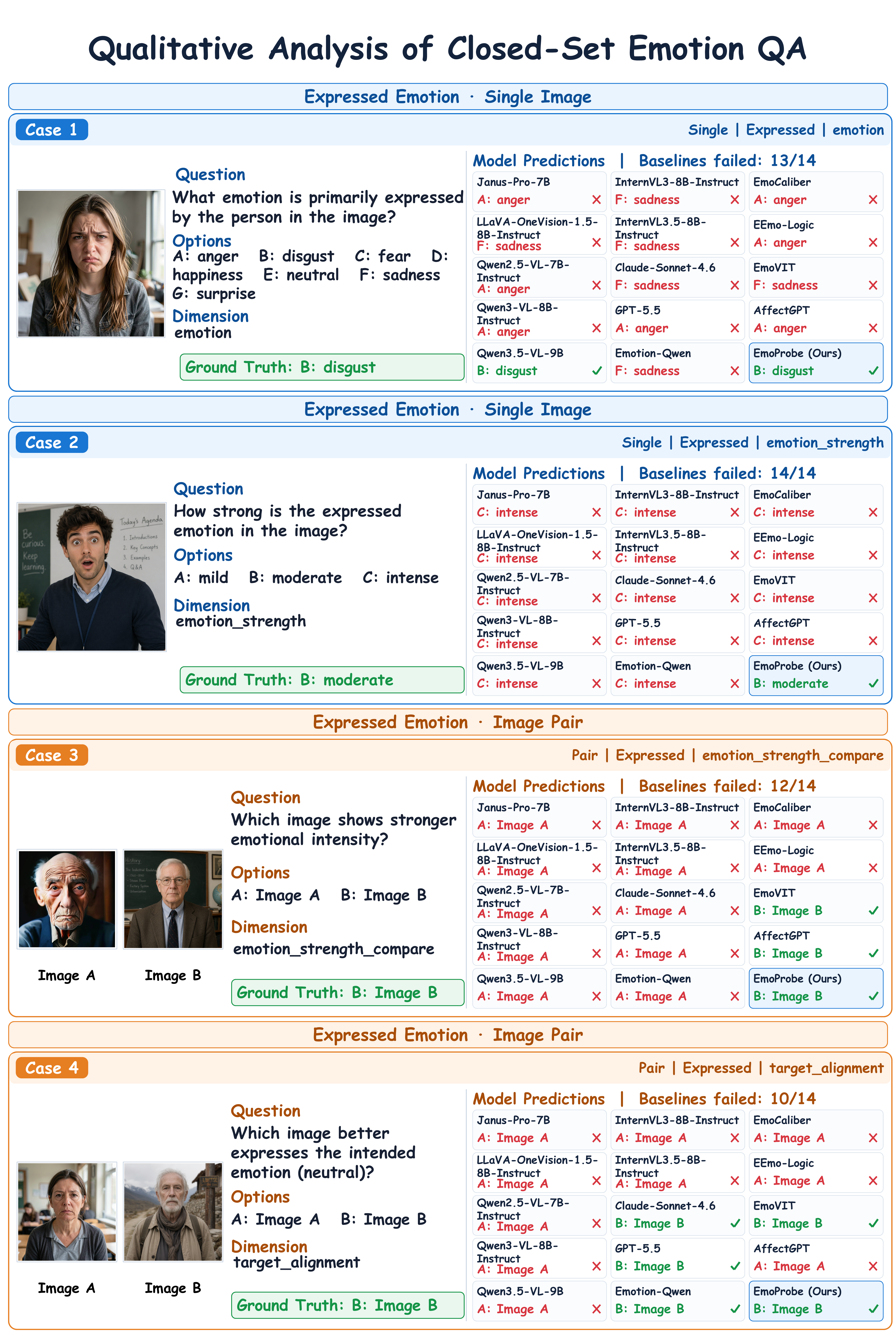}
\caption{Challenging closed-set perception cases for expressed emotion. The examples include single-image recognition of emotion category and expression intensity, together with image-pair comparisons of expression intensity and target-emotion alignment. Green and red markers indicate correct and incorrect responses, respectively; EmoProbe provides the correct response in each case.}
\label{fig:qualitative_expressed_qa}
\end{figure*}

We provide additional qualitative results that complement the aggregate evaluation in the main paper. Fig.~\ref{fig:qualitative_continuous_assessment} compares continuous predictions with human-aggregated scores for valence, arousal, dominance, and aesthetic quality, whereas Figs.~\ref{fig:qualitative_evoked_qa} and~\ref{fig:qualitative_expressed_qa} present challenging closed-set questions spanning single-image judgments and image-pair comparisons. In the illustrated cases, EmoProbe closely follows the human scores and correctly resolves questions for which most comparison models produce incorrect responses. Together, these examples show how the model supports both fine-grained continuous assessment and categorical emotion perception across the expressed- and evoked-emotion perspectives.

\end{document}